\documentclass[11pt]{article}

\usepackage[final]{acl}
\usepackage{stroop}

\usepackage{times}
\usepackage{latexsym}

\usepackage[T1]{fontenc}

\usepackage[utf8]{inputenc}

\usepackage{hyperref}

\usepackage{microtype}

\usepackage{inconsolata}

\usepackage{graphicx}

\usepackage{framed}
\usepackage{fvextra}

\usepackage[table]{xcolor}

\definecolor{mygreen}{RGB}{38,136,38}
\definecolor{myblue}{RGB}{26,101,153}
\definecolor{myorange}{RGB}{217,108,11}

\usepackage{tcolorbox}
\usepackage{xcolor}
\usepackage{fvextra}
\newtcolorbox{greenprompt}{
  colback=green!5,
  colframe=green!70!black,
  boxrule=1.2pt,
  arc=0pt,
  left=1mm,
  right=1mm,
  top=1mm,
  bottom=1mm
}

\newtcolorbox{redprompt}{
  colback=red!3,
  colframe=red!70!black,
  boxrule=1.2pt,
  arc=0pt,
  left=1mm,
  right=1mm,
  top=1mm,
  bottom=1mm
}

\title{Ladders in Chaos: When, How, (and Perhaps Why) Does Test-Time Scaling Improve LLM Machine Translation}

\author{
Di Wu$^{*,1}$ \, 
Sergey Troshin$^{*,1}$ \,
Christof Monz$^{1}$ \,
Antske Fokkens$^{2}$  \,
Vlad Niculae$^{1}$ \\
$^{1}$University of Amsterdam \\
$^{2}$Vrije Universiteit Amsterdam \\
\texttt{\{d.wu,s.troshin,c.monz,v.niculae\}@uva.nl} \\
\texttt{antske.fokkens@vu.nl}
}

\begin{document}
\maketitle
\def\thefootnote{*}\footnotetext{Equal first-author contribution.}\def\thefootnote{\arabic{footnote}}
\begin{abstract}

Two forms of test-time scaling for Large Language Models (LLMs) have emerged as effective and widely adopted paradigms: \textbf{sequential}, in which later answer attempts depend on earlier ones, and \textbf{parallel}, such as i.i.d.\ sampling with reranking. 
In this study, we investigate their properties in translation. 
First, our study shows that sequential sampling has a higher performance ceiling, providing a more diverse and effective pool of samples, particularly under smaller sampling budgets.
Second, we interrogate the nature of test-time scaling through a multidimensional manual analysis. Human analysis of the Best-of-$N$ translations demonstrates that sequential sampling substantially improves translation fluency and naturalness, but can degrade accuracy when inference budgets are large.
Finally, we suggest an explanation of the mechanism through which sequential scaling improves machine translation. Our controlled analysis partially attributes the success of sequential self-improvement to the model's access to a larger target-side context. 
Ablation experiments on sequential sampling demonstrate its robustness across different sampling temperatures, while also revealing sensitivity to context construction, suggesting directions for future improvement.

\end{abstract}

\section{Introduction}

Large Language Models (LLMs) have been shown to benefit from scaling test-time computation~\citep{snell2024scaling,muennighoff-etal-2025-s1}. In particular, iterative generation allows models to progressively refine their outputs based on previously generated responses, potentially yielding better solutions for many tasks, including mathematics and coding.

In Machine Translation (MT), LLM based systems represent the current state of the art in translation quality~\citep{kocmi-etal-2024-findings,kocmi-etal-2025-findings}, where test-time scaling technology has also recently attracted increasing attention. 
Existing studies typically demonstrate its effectiveness through complex multi-agent systems or carefully designed reasoning pipelines \cite{briakou-etal-2024-translating,wang2025maatsmultiagentautomatedtranslation}. Although promising, their complexity makes it difficult to isolate the effects of individual components, thereby limiting generalizability and hindering a deeper understanding of the underlying mechanisms. For example, \citet{wu2025please} show that simply prompting an LLM to translate again and refine can achieve even better performance than a complex, human-like reasoning pipeline.

To better understand the mechanisms underlying test-time scaling in machine translation, we study two commonly used paradigms in a maximally simplified and controlled setting: (1)~\textbf{parallel}, for which the translations are \emph{sampled} independently, followed by reranking using external metric model, \ie, Best-of-\textit{N} (BoN) sampling, which in the machine translation community is known as Quality-Aware Decoding \citep[QAD,][]{fernandes-etal-2022-quality}; and (2)~\textbf{sequential}, where translations are revisited during subsequent rounds of generation \textit{in a self-improving manner}, as exemplified by \citep{wu2025please}. Beyond machine translation, this paradigm has been interpreted as an advanced LLM capability, namely self-reflection~\citep{madaan2023selfrefine,muennighoff-etal-2025-s1}.

These two representative sampling strategies explore the decoding space in orthogonal ways, at the cost of increased computational overhead. In this study, we control the sampling budget, i.e., the number of generation rounds, to be comparable across strategies and investigate:
(1) which sampling strategy is more sample-effective and when;
(2) in which dimensions of translation evaluation would parallel and sequential be different?
(3) what mechanisms drive the performance differences; and 
(4) what is the effect of context shaping on sequential sampling.
The outcome of our analyzes can be summarized as follows:
\begin{itemize}
\item Sequential sampling explores the output spaces more effectively, compared to parallel sampling, especially under a limited budget.

\item According to human evaluation results, sequential sampling improves fluency, accuracy, and naturalness of translations; however, relative to parallel sampling, sequential can on average lose in accuracy under larger budget.

\item Our controlled experiments attribute the gains of sequential sampling in part to \textbf{its access to a larger target-side context}, rather than to the more commonly assumed self-reflection capability~\citep{madaan2023selfrefine,kamoi-etal-2024-llms,muennighoff-etal-2025-s1}.

\item We further show that simply restricting the context to a single generation round in sequential sampling leads to clear performance gains across metrics, highlighting the double-edged role of context in sequential scaling: more context can be counterproductive.
\end{itemize}

\section{Background}

\paragraph{Test-time scaling.} This strategy uses additional inference-time compute to improve model performance. 
Sequential test-time scaling in LLM research is typically realized through a combination of Chain-of-Thought ~\citep[CoT,][]{wei2022chain} and iterative multi-turn generation. The former is generally regarded as a manifestation of reasoning capability \citep{guo2025deepseek}, whereas the latter is often referred to as self-refinement~\citep{madaan2023selfrefine}. 
OpenAI’s o1 model~\citep{jaech2024openai} demonstrated such capabilities, while \citet{muennighoff-etal-2025-s1} extended the inference budget by force-generating a special ``Wait'' token, when a model tries to end generation.

Although promising, the underlying mechanisms behind this paradigm remain unclear, and its effectiveness varies considerably across domains. For example, \citet{sprague2025cot} show that CoT-style reasoning is primarily effective for math and symbolic reasoning tasks, while yielding substantially smaller gains on other types of tasks. In translation, \citet{wu2025please} demonstrate that explicitly decomposing the translation process through CoT does not lead to clear performance improvements. Instead, simply prompting LLMs to ``translate again for a better version'' yields even better results. These findings motivate our research to focus on iterative self-refinement as the main form of sequential test-time scaling.

\paragraph{Self-refinement.}

The refinement discussion for machine translation can be traced back to the automatic post-editing (APE) line of work, which has a long history in statistical machine translation \citep{knight_postediting_1994, shen-etal-2004-discriminative, simard-etal-2007-rule, correia-martins-2019-simple, fernandes-etal-2022-quality}. %
Recent refinement approaches investigate passing feedback or previous generation directly to LLMs to steer subsequent generation. These methods can generally be divided into two categories based on the source of feedback: (1) approaches that leverage external translation quality signals~\citep{huang-etal-2024-aligning,wang2025maatsmultiagentautomatedtranslation}, and (2) approaches that generate feedback internally~\citep{ki-carpuat-2024-guiding,feng-etal-2025-tear}, a process commonly referred to as self-reflection or self-refinement~\citep{madaan2023selfrefine}.

In our study, we follow~\citet{wu2025please} and consider a minimal setting for self-refinement to emphasize simplicity (\Cref{sec-3.2}). Note that we employ an external metric model for Best-of-$N$ selection solely to demonstrate the potential of the generated sample pool at evaluation time; \ie, the metric signal is never used to guide the generation process.

\paragraph{Sequential and parallel sampling.}
Our work builds on top of the somewhat tangential line of work which targets sampling strategies for diverse LLM outputs. 
Parallel (i.i.d.) or ancestral sampling techniques were developed and improved by the LLM community to trade-off between creativity and precision \citep{Holtzman2020TheCuriousCase, basu2021mirostat, hewitt-etal-2022-truncation, nguyen2024turningheatminpsampling, vilnis_emb_sampling23}. Broadly, parallel sampling and its variants are used for Quality-Aware Decoding \citep{fernandes-etal-2022-quality} or Minimal Bayes Risk decoding \citep{kumar-byrne-2004-minimum, eikema-aziz-2020-map}; to enable test-time controllability \citep{mudgal2024controlled, deng-raffel-2023-reward}, conformal generative modeling \citep{kladny2025conformal}, reasoning with diverse decoding paths \citep{wang2024mmlupro}; and for ambiguity resolution \citep{kobalczyk2025active, chen2025learning, saparina-lapata-2025-disambiguate}.

Sequential (non-i.i.d.) sampling in its various forms is often used as a way to improve sampling efficiency. \citet{ilia-aziz-2024-predict} prompt API models to enumerate responses; \citet{saparina2024ambrosia} prompt language models to enumerate possible interpretations of ambiguous questions; \citet{troshin-etal-2025-asking} and \citet{ zhang2025verbalizedsamplingmitigatemode} formulate variants of sequential sampling to generate diverse outputs from LLMs. 

\section{Sequential and Parallel Sampling}
In this section, we formally define parallel and sequential sampling. Both sampling algorithms can be applied up to a fixed number of generation rounds (budget), ensuring fair comparison.

\subsection{Parallel Sampling}
\label{sec-2.1}
Given a language model $p_{\theta}$ and a source text $x$ to be translated, the output can be viewed as a sample $y \sim p_\theta(\cdot \mid I(x))$, where $I(x)$ is a prompt that includes $x$ as a component to request translation. In contrast to beam-search, parallel sampling draws \textbf{translations} independently $t$ times from a LLM conditioned on the same input prompt:
\begin{equation}
    y^{(i)} \sim p_{\theta} (\cdot | I(x))~~\text{for $i \in \{1,\ldots, t\}$ };
\end{equation}
In Appendix~\ref{app-1.1}, we list the prompt format we used for our experiments. We do not use any form of prompt tuning, but we explicitly specify output format expectations.

\subsection{Sequential Sampling}
\label{sec-3.2}
Sequential sampling for translation is a multi-turn process in which the \textbf{translations} of previous steps serve as context for the next generation turn. For a turn $t$, \textit{sequential} samples translations $y$ as follows:

\begin{equation}
    y^{(t)} \sim p_\theta\left(\cdot \mid I^{(t-1)},y^{(t-1)}, ..., I^{(1)},y^{(1)}, I(x)\right),
\end{equation}

\noindent where $I^{(\cdot)}$ is a prompt requesting a model to revisit a translation to improve it \citep{wu2025please}. We use the same prompt for each revisiting turn, \ie, $I^{(\cdot)} := I$. The revisiting prompt $I$ in our case is:
\begin{center}
\fbox{\parbox{0.9\linewidth}{
``Please translate again for a better version.''
}}
\end{center}

\begin{figure*}[t]
    \centering
    \includegraphics[width=0.95\textwidth]{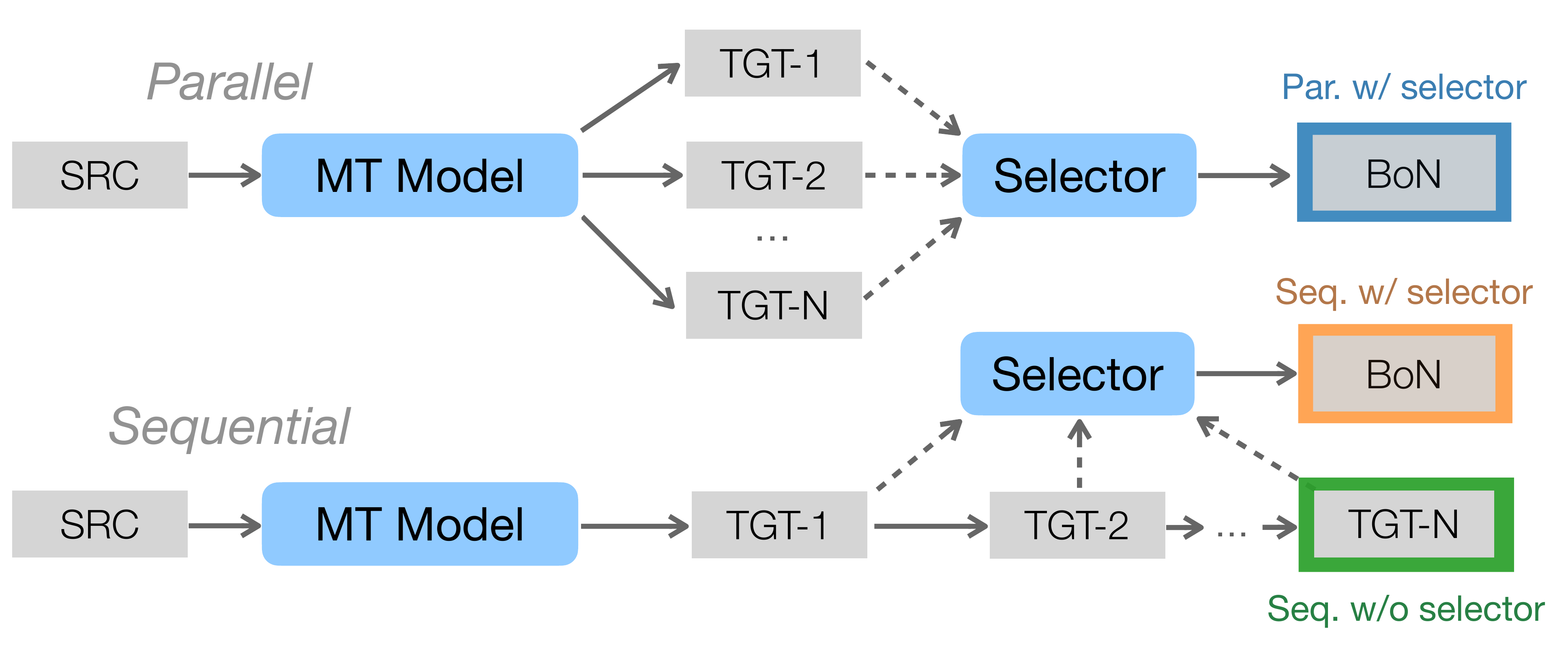}
    \caption{Illustration of the three sampling strategies investigated in this paper: 1) parallel sampling with selector, commonly referred to as Best-of-$N$ sampling; 2) sequential sampling without a selector (sequential self-refinement), where we follow \citet{wu2025please} to prompt model to \textit{translate again for a better version}, and 3) sequential sampling with selector, which enables a fair comparison with parallel sampling (Best-of-$N$ sampling).}
    \label{fig:fig-2}
\end{figure*}

All sequential sampling strategies in this work use this prompt, 
because (1) it provides a maximally simplified setting, (2) it emphasizes the quality requirement, and (3) \citet{wu2025please} show that this minimal prompt achieves the best results on the WMT24 dataset~\citep{kocmi-etal-2024-findings}, outperforming carefully designed step-by-step prompting methods~\citep{briakou-etal-2024-translating}.

Sequential sampling follows the test-time scaling framework in LLMs research: multiple rounds of translation are introduced at inference time, while explicitly requiring the model to produce better translations, analogous to test-time reflection ability in LLMs~\citep{madaan2023selfrefine,muennighoff-etal-2025-s1}. Notably, this setting removes the explicit chain-of-thought (CoT) process, as recent studies have shown that explicit CoT provides no clear benefit for general machine translation~\citep{wu2025please,rajaee2026unlockingreasoningcapabilitymachine}.

\subsection{Sampling with a Selector}
The two sampling strategies explore the decoding space in different ways. Our study investigates their respective properties in terms of randomness and potential of performance. However, investigating the performance scaling is nontrivial, as it requires identifying the best outputs from a large pool of stochastic generations. 
To this end, we follow a simple instance of Quality-Aware Decoding (QAD) strategy~\citep{fernandes-etal-2022-quality} and employ an external reference-free metric model~\citep{rei-etal-2023-scaling} as the selector. This setting is often referred to as Best-of-$N$ sampling in LLMs research. 

More specifically, for a batch of translation $y_{p}^{1...N}$ and $y_{s}^{1...N}$, generated by \textit{parallel} and \textit{sequential} sampling given the same input $x$, we investigate the diversity and performance under the following three settings:

\begin{enumerate}
    \item \textbf{Parallel w/ selector}: we independently sample $N$ outputs for each source sentence, then use a reference-free metric model to select and report the corresponding best performance.
    \item \textbf{Sequential w/o selector}: we sequentially sample $N$ outputs for each source sentence, then directly report average translation performance in each sampling step.
    \item \textbf{Sequential w/ selector}: we sequentially sample $N$ outputs for each source sentence, then use a reference-free metric model to select and report the corresponding best performance.
\end{enumerate}

Figure~\ref{fig:fig-2} exemplifies the three decoding strategies. 
In this work, we vary the sampling budget for each source sentence from 1 to $N$ across the three settings to illustrate the variation in properties and performances. In the following study, we sample under the temperature of 1.0, except for temperature ablations (see Appendix~\ref{app:sampling_params}). We discuss other generation details in Appendix~\ref{app:generation_details}.

\begin{figure*}[t]
    \centering
    \setlength{\tabcolsep}{2pt}

     \includegraphics[width=1.0\textwidth]{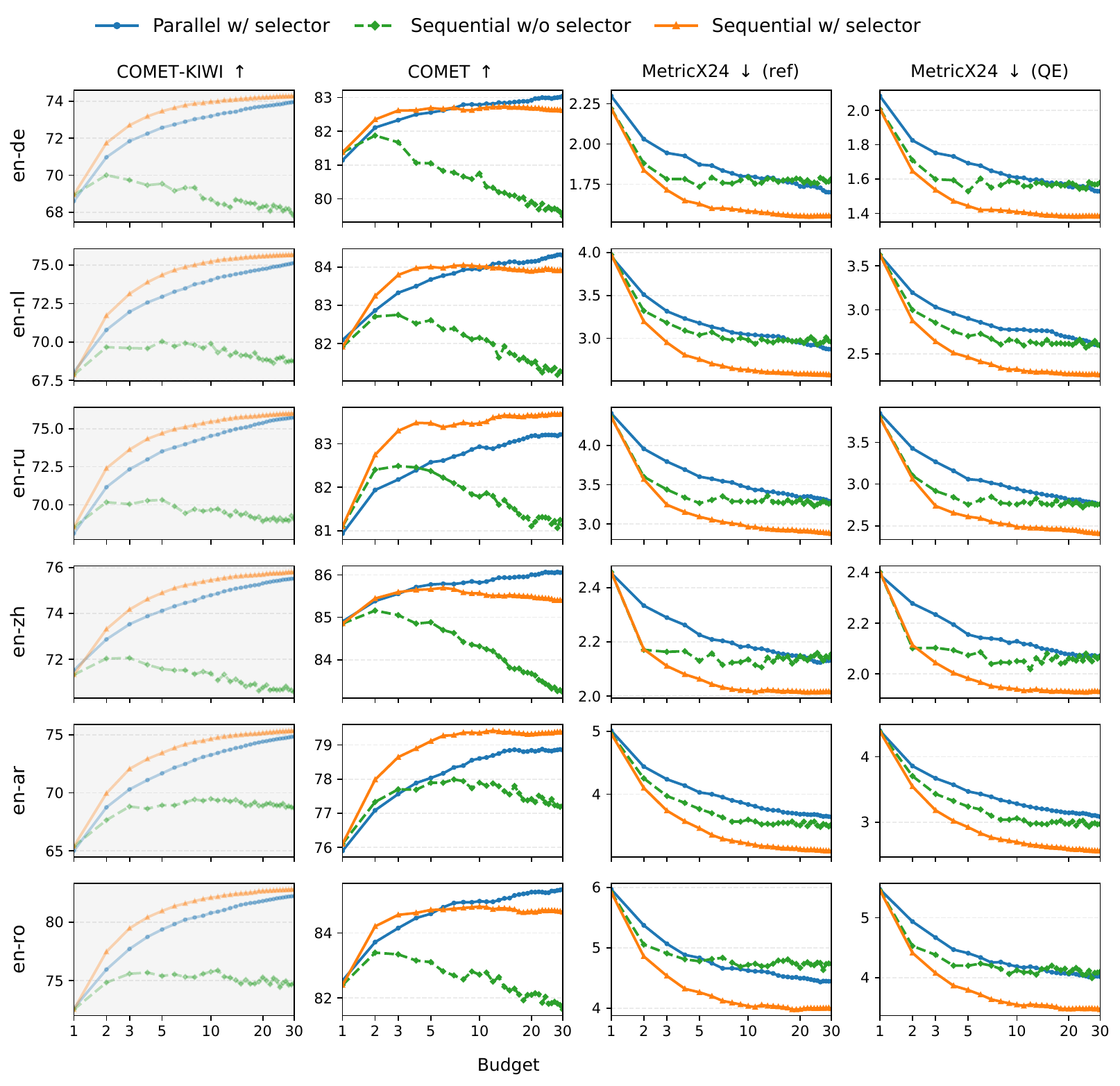}

    \caption{MT quality vs.\ the number of rounds of translation (budget) for the \texttt{Qwen-3-32B} model.}
    \label{fig:main_plots_32b_model}
\end{figure*}

\section{Experimental Setup}
\subsection{Dataset}
We use the WMT24++ dataset \citep{deutsch-etal-2025-wmt24}~\footnote{Extended version of the WMT24 \citep{kocmi-etal-2024-findings}.}, intended for evaluation of MT systems, as our test set, because 1) it covers a variety of domains and 2) the translation references in WMT24++ are
human-written and subsequently post-edited by professional translators, ensuring the highest possible data quality. We filter out sentence pairs that are marked as low-quality or invalid using `is\_bad\_source' labeled field, following WMT24++ recommendations~\footnote{This leaves $960$ segments for each language.}. To maximize the reliability of the metric models we used, we conduct experiments on a subset of 6 well-supported high-- and medium-resource language directions, \ie, \texttt{en$\rightarrow$zh,de,ru,nl,ro,ar}\footnote{zh: \texttt{zh\_CN}, ar: \texttt{ar\_SA}}, representing different writing systems and language families. %

\subsection{Models}
We experiment with capable open-source instruction-tuned models from the Qwen3 family across different parameter scales \citep{yang2025qwen3technicalreport}, namely \texttt{Qwen3-32B} (instruct mode) and \texttt{Qwen3-4B-Instruct-2507} (referred to as \texttt{Qwen3-4B}); 
Following \citet{wu2025please}, we also conduct experiments with a commercial LLM \texttt{GPT-4o-mini}\footnote{gpt-4o-mini-2024-07-18}.

\subsection{Metrics}
\label{sec:3-3}
Unless otherwise specified, we use \href{https://huggingface.co/Unbabel/wmt23-cometkiwi-da-xl}{CometKiwi-XL} \citep{rei-etal-2023-scaling}, the official reference-free metric for WMT24~\citep{kocmi-etal-2024-findings} evaluation, as the selector for the following experiments. 
Meanwhile, to maximize the difference compared to the selector in case of  excessive metric hacking~\citep{NEURIPS2022_3d719fee,kovacs-etal-2024-mitigating,NEURIPS2025_e46f7f5c}, we report translation quality in a few widely used metric models, including (i) MetricX-24 \citep{juraska-etal-2024-metricx}, a hybrid metric used in the QE and reference-based modes, namely \href{https://huggingface.co/google/metricx-24-hybrid-xxl-v2p6-bfloat16}{MetricX-24-Hybrid-XXL (qe)} and  \href{https://huggingface.co/google/metricx-24-hybrid-xxl-v2p6-bfloat16}{MetricX-24-Hybrid-XXL (ref)}; (ii) a reference-based \href{https://huggingface.co/Unbabel/wmt22-comet-da}{COMET} \citep{rei-etal-2022-comet}; (iii) a reward model \href{https://huggingface.co/ShaomuTan/ReMedy-9B-22}{ReMedy-9B-22} \citep{tan-monz-2025-remedy},
used in the QE and reference-based mode. For the main results, we plot a translation metric of a selected sample (Best-of-$N$, or the last output of the self-refinement strategy) at a given budget level, where budget denotes the number of translation rounds.

\section{An Overview of Sampling Effectiveness}
\label{sec:5}
In this section, we first give an overview of the effectiveness across different sampling strategies and evaluation metrics, under progressively increasing computation budget. Figure~\ref{fig:main_plots_32b_model} presents the results for test-time scaling under multiple settings. \Cref{tab:main_results} presented aggregated results for all 3 models, aggregated over the language directions. As described in \Cref{sec:3-3}, we use \texttt{CometKiwi-XL} as a selector. We summarize the key findings as follows.

\begin{figure*}[t]
\centering
\smaller

\includegraphics[width=1.0\textwidth]{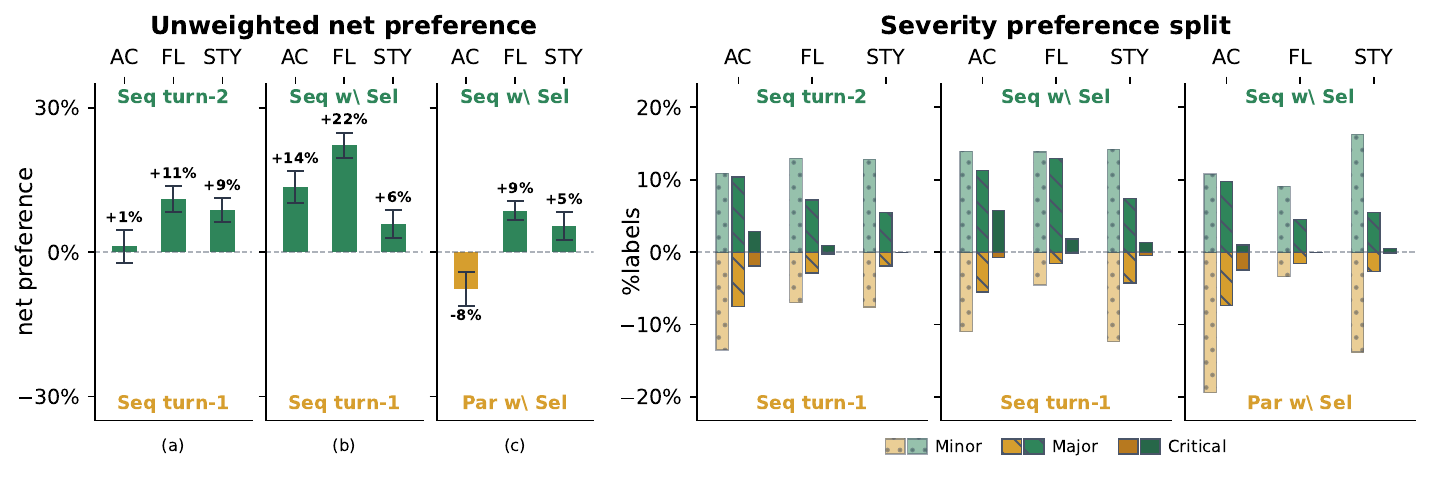}

\caption{Human contrastive evaluation results: averaged preference split across three MQM quality dimensions: \textbf{accuracy}~(AC), \textbf{fluency}~(FL) and \textbf{style}~(STY) for the \texttt{Qwen3-32B} model, averaged over $5$ languages. (Left) shows the proportion of unweighted net preference (in \%), and (right) presents preferences split across severity levels. Contrastive comparisons are (a) sequential turn-$1$ vs turn-$2$, (b) sequential with selector (budget-$30$) vs turn-$1$, (c) sequential vs parallel (with selector, budget-$30$). Error bars represent standard error of the mean (mean $\pm$ 1SE). \label{fig:human_annotation_plot_full}}

\end{figure*}

\paragraph{The self-refinement strategy works better in the early rounds.} 
The \textcolor{mygreen}{green line} in Figure~\ref{fig:main_plots_32b_model} denotes the performance of self-refinement strategy in different metrics. It is clear that within the first few rounds (\eg, 2 or 3), there are substantial and consistent performance improvements over the first round across all language directions and evaluation metrics. For example, for \texttt{en$\rightarrow$ru} direction, the improvements reach approximately $1.5$ COMET, $1.0$ MetricX24-ref, and $1.5$ MetricX24-qe points when employing the second round of translation. Our human evaluation aligns well with this finding (\Cref{fig:human_annotation_plot_full} left-a). These findings are also aligned with \citet{wu2025please}, where they report quality improvements for earlier rounds. Self-refinement patterns persist for commercial LLMs as well (\Cref{tab:main_results}, \Cref{fig:gpt_4o_mini}).
Recent LLMs research attributes these improvements to LLMs' reflection ability~\citep{madaan2023selfrefine,muennighoff-etal-2025-s1}. In \Cref{sec:6-2}, we revisit this phenomenon and provide a more concrete explanation.
When we evaluate the quality of sequential sampling for later rounds of translation (\eg, $>$ 5 rounds), the performance quickly plateaus and may even decline in some of the languages and metrics. In \Cref{sec:6-3}, we analyze the output space and partially attribute the plateau effect to the reduced marginal diversity.

\paragraph{Using selector is highly effective.} 

The \textcolor{myblue}{blue} and \textcolor{myorange}{orange} lines in \Cref{fig:main_plots_32b_model} represent the results of parallel and sequential sampling with an external selector under different computation budgets, respectively. Notably, both methods substantially outperform direct translation (\ie, budget of $1$), even under very limited budgets (\eg, $2$). Increasing the sampling size further improves performance across most metrics and language directions, with gains becoming substantial and beginning to converge for budget-$30$. The first column (in gray) in \Cref{fig:main_plots_32b_model} corresponds to the setting in which CometKiwi-XL is used as both the selector and the evaluator,\footnote{A setting most prone to metric hacking.} shown here only to profile the Best-of-$N$ sampling efficiency.

Although the early iterations of self-refinement clearly improve average translation quality, the quality can degrade as the number of samples increases (at least according to COMET).\footnote{%
Our findings match the observation that
COMET tends to 
emphasize
adequacy over fluency \citep{park-pado-2024-multi}.} 

By comparing results for sequential with and without selector (\textcolor{myorange}{orange} vs \textcolor{mygreen}{green}), we observe that adding a selector results in better samples, given higher compute budgets. This suggests that the \textbf{language model alone cannot consistently produce better samples given more budget rounds}, and highlights the need for a selector.
We thus treat sequential sampling without selector as a more stochastic and low-trust process.

\paragraph{Sequential sampling explores the decoding space more effectively,} especially under relatively lower sampling budgets.
Automatic evaluation results show that sequential sampling outperforms parallel sampling for smaller computational budgets (in all metrics), but sometimes loses to parallel for higher compute budgets (for the COMET metric); for the MetricX-24 metric, sequential always outperforms parallel. 
Our main finding persists for commercial LLMs and for a smaller Qwen3 model (\Cref{tab:main_results}).
Overall results are very promising for sequential sampling, but slightly contradictory \wrt the performance ceiling, given the difference over metrics.
To understand the results better, we conduct human annotation (\Cref{sec:6-1}) to inspect the improvements targeting not a single quality score, but 3 distinct MQM dimensions.

\section{Analysis}
\begin{figure*}[t]
    \centering
    \includegraphics[width=0.98\textwidth]{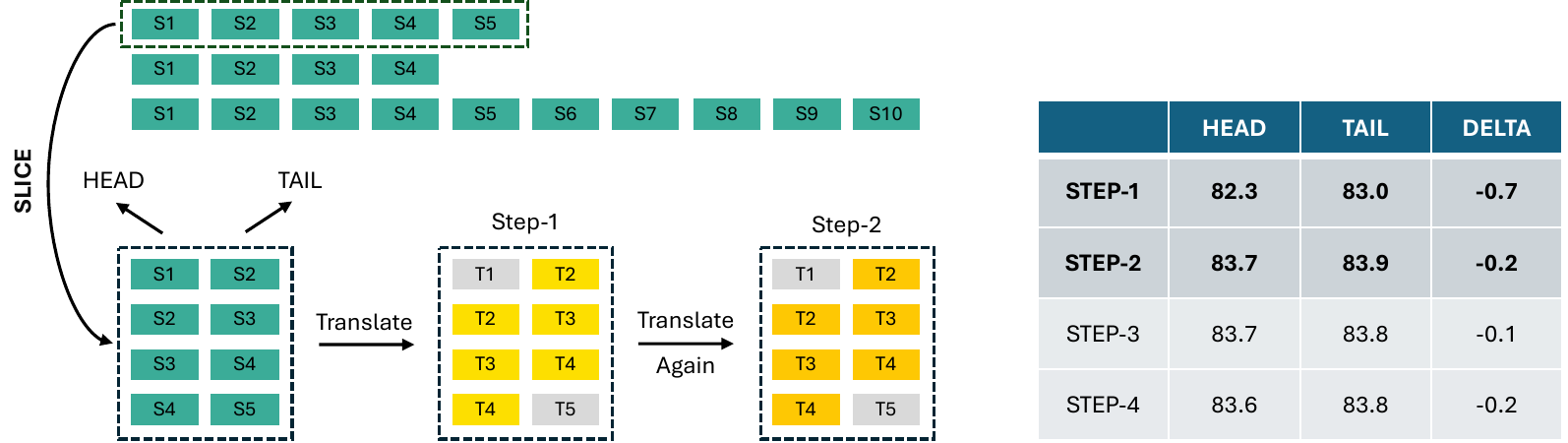}
    \caption{Illustration of the controlled experiments in \Cref{sec:6-2}. Each document in WMT24++ is sliced into a batch of two consecutive sentences and translated using the self-refinement strategy. We compare the average translation performance at the HEAD and TAIL positions and report the results in the table on the right. It is clear that Step-2 recovers performance for the HEAD position. Note that we discard $T_1$ and $T_5$ to ensure that each sentence appears exactly once as a HEAD segment and once as a TAIL segment, respectively, thereby enabling strict comparability.}
    \label{fig:fig-3}
\end{figure*}

\subsection{Human Evaluation}
\label{sec:6-1}
To better understand the automatic results for sequential vs parallel, we ask annotators to compare translations under three different settings: (a) sequential turn-1 vs turn-2, (b) sequential with selector vs turn-1, and (c) sequential vs parallel sampling with selector. We focus on three MQM dimensions~\citep{freitag-etal-2021-experts}: accuracy (AC), fluency (FL) and style (STY), see \Cref{app:annotation} for detailed annotation policy. 
Results are presented in \Cref{fig:human_annotation_plot_full} and categorized into ``Unweighted preference'' (left) and ``Severity split'' (right) versions. The former figure  represents the net ratio of winning translations (average) between a pair of compared methods, while the latter denotes the proportion of annotations for each dimension split by their error severity (minor, major, critical), where green or orange marks the direction (win/lose).

Compared with direct translation, sequential sampling yields clear improvements across all dimensions with just one additional round in \Cref{fig:human_annotation_plot_full}~left-(a).
This performance gap is further amplified when combined with a selector under higher budgets as can be seen in \Cref{fig:human_annotation_plot_full}~left-(b). When we look at the proportion of wins by severity (\Cref{fig:human_annotation_plot_full}~right), we notice a few critical improvements, most of which are attributed to sequential.

In \Cref{fig:human_annotation_plot_full}~left-(c), when we compare the selected samples from sequential vs parallel under higher budgets (30), we observe that sequential generates more fluent outputs with better style, but samples from parallel are more accurate. A possible explanation for the lower accuracy of the sequential approach is that, as multi-turn generation progresses, the influence of the source sentence gradually weakens, leading to decreasing faithfulness to the source. Consistent with this hypothesis, our ablation results (\cref{sec:ablation}) demonstrate that when the growing context is cut, sequential sampling recovers more stable quality behavior (according to COMET). More detailed annotation results are presented in \Cref{fig:human_annotation_plot_full-2}, revealing substantial cross-lingual variation across languages. In the English-to-Dutch direction, annotators judged the outputs to be of lower quality, whereas in the English-to-Chinese direction, annotators reported high quality of the outputs. These differences can plausibly be attributed to variations in model performance across languages. 
Our manual review of the samples also suggests that sequential rephrasing can be quite creative, as models embrace the variability of natural language, which might pose challenges for selector models.

\subsection{On the Success of Self-Refinement}
\label{sec:6-2}
As shown in \Cref{fig:main_plots_32b_model}, sequential sampling is particularly effective in the early steps. A key question is what drives these improvements. Previous studies in the LLM community attribute this to the ``reflection'' abilities~\citep{madaan2023selfrefine,guo2025deepseek,muennighoff-etal-2025-s1}, although the underlying mechanism remains unclear and is often vaguely ascribed to human-like ``intelligence''. 
Here, we have a more concrete explanation.

Language model generation follows an auto-regressive, left-to-right process, which naturally induces an asymmetric context structure: within a span of generation, later generation steps have seen the previous translated outputs to their left, but the earlier generation steps have no access to their right context. This phenomenon has long been studied in context-aware/document translation~\citep{voita-etal-2018-context,voita-etal-2019-context,voita-etal-2019-good,lopes-etal-2020-document}. 

We argue that part of the success of self-refinement stems from the same effect: during first-pass generation, earlier and later segments differ in quality due to an asymmetric context. In contrast, the second-pass translation has access to the full global context from the first pass, allowing it to recover and improve translation quality. To demonstrate and isolate the effect of the asymmetric context, we conduct a controlled experiment, presented in \Cref{fig:fig-3}:

\begin{enumerate}
    \item We split each document into consecutive sentences using a sliding window of size~$2$ to form a dataset such that each sentence appears \emph{exactly once} as a HEAD segment and \emph{exactly once} as a TAIL segment\footnote{As shown in \Cref{fig:fig-3}, we remove $T_1$ in the head position and $T_5$ in the tail position to strictly ensure this constraint.}, ensuring strict comparability across positions. In practice, we use WMT24++ and maintain the sentence-to-document mapping using the metadata.
    \item We then apply the sequential self-refinement strategy to the resulting corpus and compare the average translation performance between the HEAD and TAIL positions.
\end{enumerate}
We present the average COMET score results for \texttt{en-zh,ru,nl} in \Cref{fig:fig-3}, which show that: (1) Translations in the HEAD position clearly underperform their counterparts in the TAIL position during the first round, highlighting the performance gap introduced by asymmetric context; and (2) Translation quality in the HEAD position is improved and recovered to a comparable level to that of the TAIL position in the second round. More experimental details can be found in \Cref{app:D}.

These results are closely aligned with the preceding hypothesis of asymmetric context, highlighting the inherent limitations of left-to-right generation. Furthermore, they provide a concrete explanation for at least part of the improvements observed in self-refinement under sequential scaling.

\subsection{Sequential Sampling is More Diverse for Smaller Budgets}
\label{sec:6-3}
\begin{figure}[th]
    \centering
    \setlength{\tabcolsep}{2pt}

     \includegraphics[width=0.5\textwidth]{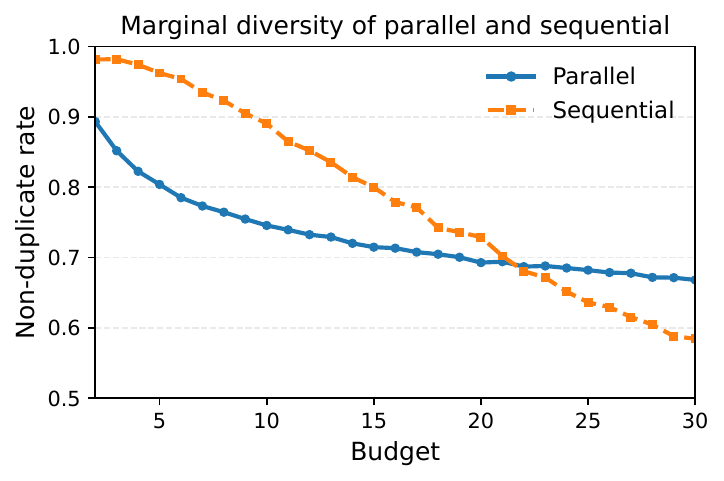}

    \caption{Diversity vs.\ budget for \texttt{Qwen-3-32B} model, avg. over 6 languages. Non-duplicate rate for each budget is computed \wrt previously sampled outputs, converging to $0$ in the budget limit. Error bars, representing standard error of the mean ($\pm 1SE$), are nearly indistinguishable.}
    \label{fig:diversity_plots_32b_model}
\end{figure}

To better understand the plateau effect of sequential sampling for larger sample budgets, we conduct an output space diversity analysis, measuring the sample-effectiveness of the later rounds of generation.
To compare the differences in output variability, we estimate a rate of marginally novel samples. When adding one sample  $y$ at a time, given a history of samples $H = \{y^1, \ldots, y^i\}$, we compare the new sample to each sample in the context pool. We then define an exact match distance as $D(y | H)~=~\underset{y^i\in H}{\min} [y \ne y^i]$,
where $D(y|H) = 0 \iff$ the new output matches at least one of the existing samples in the context and $D(y|H) = 1$ means the new output is ``novel''. For each budget value, we report the $D(y|H)$, averaged over the dataset.
Our results in \Cref{fig:diversity_plots_32b_model} show that for earlier rounds, sequential sampling generates much fewer exact matches, starting from nearly zero duplicate rate for budget-$2$, whereas parallel already starts with a $\sim\!0.9$ duplicate rate. 
For the later rounds, sequential sampling appears to produce more duplicates and show signs of the repetition bias: it starts to repeat the previous generations more often than parallel. In \Cref{sec:ablation}, we revisit this effect, showing that by controlling the context shaping, we can reduce the repetition bias.

\section{Ablation study \label{sec:ablation}}
\paragraph{Not all models benefit equally well from reflection.}
To complement the results with \texttt{Qwen-3-32B}, we run the experiments for a smaller model from the same family. \Cref{tab:main_results} (with full results in  \Cref{fig:main_plots_4b_model}) presents the results for \texttt{Qwen-3-4B}. 
We observe a similar overall pattern, \ie, \textit{sequential} achieves better performance under smaller budgets. However, its advantage over \textit{parallel} is less pronounced than in larger models, which might indicate that small models have worse ability for self-refinement.

\paragraph{Larger sample context does not imply better performance.}

\begin{table*}[t]
\centering

\begin{tabular}{@{}llrrr@{\hspace{1.2em}}rrr@{\hspace{1.2em}}>{\columncolor[RGB]{255,255,255}[\tabcolsep][0pt]}c@{}}
\toprule
 &  & \multicolumn{3}{c}{Seq.\ w/o\ selector} & \multicolumn{3}{c}{Seq.\ w/\ selector} & \multicolumn{1}{c}{Parallel} \\
Model     & Metric          &                             $t=1$ &                             $t=2$ &                             $t=3$ &                        $h=\infty$ &                             $h=5$ &                             $h=1$ &     \\
\midrule
Qwen3     & COMET           & \cellcolor[RGB]{255,255,255}81.33 & \cellcolor[RGB]{243,250,241}82.11 & \cellcolor[RGB]{242,250,240}82.13 & \cellcolor[RGB]{224,243,219}83.28 & \cellcolor[RGB]{224,243,219}83.27 & \cellcolor[RGB]{217,240,211}83.72 & \cellcolor[RGB]{221,242,216}83.47 \\
(32B)     & metricx24 (ref) &  \cellcolor[RGB]{255,255,255}3.98 &  \cellcolor[RGB]{239,249,237}3.40 &  \cellcolor[RGB]{236,247,232}3.25 &  \cellcolor[RGB]{220,241,215}2.69 &  \cellcolor[RGB]{218,241,213}2.61 &  \cellcolor[RGB]{217,240,211}2.56 &  \cellcolor[RGB]{229,245,225}3.01 \\
          & metricx24 (qe)  &  \cellcolor[RGB]{255,255,255}3.61 &  \cellcolor[RGB]{240,249,237}3.05 &  \cellcolor[RGB]{235,247,232}2.90 &  \cellcolor[RGB]{220,241,215}2.34 &  \cellcolor[RGB]{218,241,213}2.27 &  \cellcolor[RGB]{217,240,211}2.22 &  \cellcolor[RGB]{229,245,225}2.68 \\
\midrule
Qwen3     & COMET           & \cellcolor[RGB]{255,255,255}78.42 & \cellcolor[RGB]{251,254,251}78.71 & \cellcolor[RGB]{249,253,249}78.87 & \cellcolor[RGB]{225,243,220}80.86 & \cellcolor[RGB]{223,242,218}80.96 & \cellcolor[RGB]{218,240,212}81.42 & \cellcolor[RGB]{217,240,211}81.46 \\
(4B)      & metricx24 (ref) &  \cellcolor[RGB]{255,255,255}5.69 &  \cellcolor[RGB]{242,250,239}5.04 &  \cellcolor[RGB]{238,248,235}4.85 &  \cellcolor[RGB]{222,242,216}4.07 &  \cellcolor[RGB]{221,241,215}4.02 &  \cellcolor[RGB]{217,240,211}3.84 &  \cellcolor[RGB]{224,243,219}4.16 \\
          & metricx24 (qe)  &  \cellcolor[RGB]{255,255,255}5.10 &  \cellcolor[RGB]{241,249,239}4.44 &  \cellcolor[RGB]{237,248,235}4.26 &  \cellcolor[RGB]{221,242,216}3.50 &  \cellcolor[RGB]{221,241,215}3.46 &  \cellcolor[RGB]{217,240,211}3.29 &  \cellcolor[RGB]{224,243,219}3.63 \\
\midrule
GPT       & COMET           & \cellcolor[RGB]{255,255,255}82.22 & \cellcolor[RGB]{243,250,241}82.86 & \cellcolor[RGB]{242,250,240}82.93 & \cellcolor[RGB]{222,242,217}83.96 & \cellcolor[RGB]{221,242,216}84.01 & \cellcolor[RGB]{217,240,211}84.23 & \cellcolor[RGB]{225,243,220}83.81 \\
(4o-mini) & metricx24 (ref) &  \cellcolor[RGB]{255,255,255}3.56 &  \cellcolor[RGB]{240,249,238}3.13 &  \cellcolor[RGB]{235,247,232}3.00 &  \cellcolor[RGB]{218,240,212}2.51 &  \cellcolor[RGB]{217,240,211}2.48 &  \cellcolor[RGB]{218,240,212}2.50 &  \cellcolor[RGB]{231,245,227}2.86 \\
          & metricx24 (qe)  &  \cellcolor[RGB]{255,255,255}3.29 &  \cellcolor[RGB]{240,249,238}2.88 &  \cellcolor[RGB]{236,247,233}2.74 &  \cellcolor[RGB]{218,241,213}2.25 &  \cellcolor[RGB]{217,240,211}2.21 &  \cellcolor[RGB]{218,240,212}2.24 &  \cellcolor[RGB]{231,245,227}2.60 \\
\bottomrule
\end{tabular}

\caption{Automatic evaluation results for sequential and parallel sampling strategies aggregated across 6 languages. Left side presents results for self-refinement i.e. turn-$1,2,3,$ of sequential sampling; right side shows the results with Best-of-$N$ (budget $30$) for the sequential (full context, $h=5$, $h=1$); and parallel with selector. \label{tab:main_results}}
\end{table*}

\begin{figure*}[t]
    \centering
    \setlength{\tabcolsep}{2pt}

     \includegraphics[width=1\textwidth]{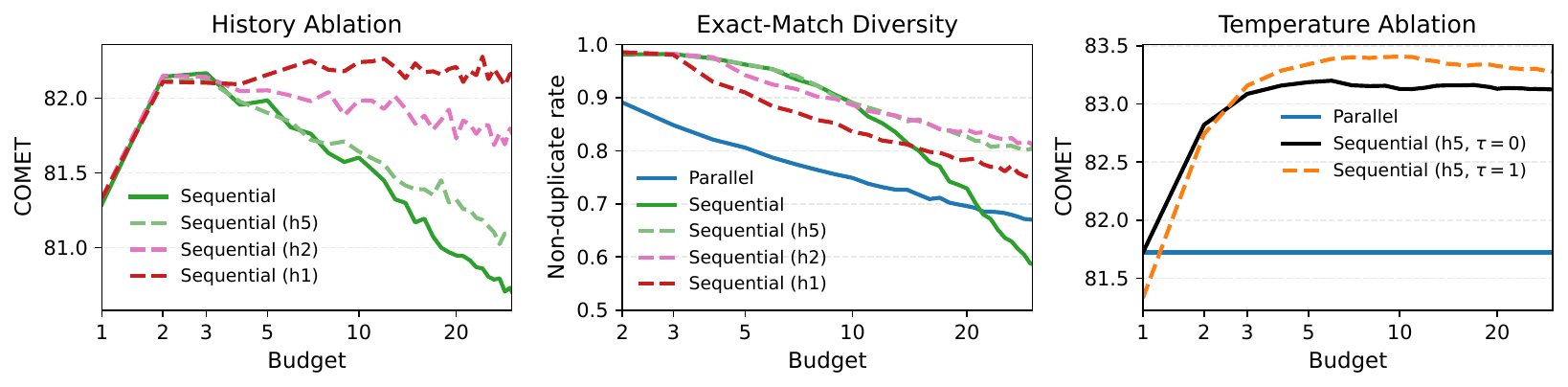}

    \caption{Ablation results for \texttt{Qwen-3-32B}. Left: equipping seq. w/o selector with smaller window size $h$ results in less COMET degradation.   Middle: context-controlled variants of sequential sampling show less diversity collapse. Right: sequential w/ selector sampling works even in the low-temperature limit, whereas parallel sampling collapses.}
    \label{fig:ablation_plots}
\end{figure*}

Our results in \Cref{fig:main_plots_32b_model} and analysis in  \Cref{sec:6-3} suggest that too long contexts of sequential sampling may lead to diminishing quality gains and repetition bias.
To study the context effect further, we propose a sequential setting by capping the maximal number of samples that we store in context, such that for any sampling round, the number of samples stored in context is bound from above. In particular, we use a simple \emph{sliding window} approach, setting $h$ as the window size parameter:
\begin{equation}
    y^{(t)}_{h}\!\sim\!p_{\theta} \left(\cdot | I, y^{(t-1)}, \ldots, y^{(t-h)}, I, y^{(1)}, I(x) \right).
\end{equation}
To better align our experiments with parallel sampling, we always keep the initial translation $y^{(1)}$, and slide over the next $h$ stored translations.

\Cref{tab:main_results} results show that perhaps counterintuitively, larger context sizes do not imply better performance. The results for sequential with constrained context are better for almost every metric and language direction. Moreover, smaller context sizes $h$ lead to less COMET degradation for the self-refinement mode  (\Cref{fig:ablation_plots}, left). From a sampling effectiveness angle, this result is particularly positive, as a smaller sequential context leads to both higher gains and lower costs. To explain this effect, we extend the diversity analysis (\Cref{sec:6-3}) to the context-size ablation experiment. From \Cref{fig:ablation_plots} (middle), we observe that controlling the context effectively reduces the repetition bias, which explains the better results of sequential with a controlled context. It might be promising to further study the context bias in the context of  dynamical systems, looking for cycles or attractors \citep{wang-etal-2025-unveiling}.

\paragraph{Sequential sampling does not collapse in the low temperature limit.}

When decreasing the temperature of the parallel sampling, eventually all diversity is expected to vanish, when approaching the greedy decoding. Sequential sampling, however, has the potential to remain useful for Best-of-$N$ sampling even under extremely low temperature mode (including greedy decoding). \Cref{fig:ablation_plots} (right) confirms this hypothesis, showing the results for sequential with selector comparing greedy sampling vs high-temperature sampling. This observation may be useful for future work on developing better sampling methods, that optimize the quality-diversity trade-offs.

\section{Conclusion}
We conducted a set of experiments on open-source and closed-source models to compare two widely used paradigms for test-time scaling: sequential and parallel. Our results highlight the superiority of the sequential paradigm in terms of sample efficiency and low output-space duplicate rate. Our analysis and ablation study partially explain the success of sequential sampling through the lens of context processing; we further underline the importance of context control, demonstrating more stable sequential sampling under context-constrained scenarios, paving a road for further improvement.

\section{Limitations}

One limitation of our work is that we test against a single prompt format, even though our prompting strategy is motivated by previous work \citep{wu2025please}. We expect that our findings can change, if one replaces the ``Please translate again for a better version'' prompt with other types of refinement instructions, for example purely diversity focused. 

Additionally, we observe that the sequential reflection abilities of different models can vary, thus our findings may or may not generalize to other models, depending on the model's capabilities. 

Furthermore, sequential sampling may produce more fluent and natural translations, containing more minor inaccuracies, which can be risky in practical MT settings because fluent but unfaithful translations are hard for users to detect.

Another limitation is that we tested our approach using a single dataset and only a small subset of languages. Using only a single dataset is somewhat balanced out given that WMT24++ covers a broad range of domains and language styles. We didn't test on low-resource or extremely low-resource languages, prioritizing covering a set of languages, for which we have access to human annotators. Our MQM analysis only targets $3$ directions of translation quality, and does not cover other directions such as terminology use or locale conventions. For every experiment, due to resource constraints, we only sampled a single trajectory for each sequential run, which limits the stability of the reported quality values. 

Finally, our findings regarding the ``Why'' questions are limited to the particular methodological strategies we used to monitor the performance differences. Our diversity analysis is limited to hard matches (duplicate rate), and can be extended using soft-matching utility functions. Our context size control analysis only suggests a single strategy to compact the context, namely sliding window approach.

\section{Ethical considerations}
We acknowledge that all generation settings discussed in this work can be biased towards or against certain attributes, can systematically delete or suppress certain aspects of translations. In particular, our results under Best-of-$N$ setup inherit the biases of the selector metric, and are shaped by what kind of translations are picked up by the selector metric. Even if the selector metric would be perfectly unbiased, in the setting of limited context size, it is possible that the LLM has biases, steering its improvements away from certain aspects. 

\section{Acknowledgments}
This work is supported in part by project VI.Veni.212.228 of the research program
`Veni', which is financed by the Dutch Research Council (NWO); this research is also funded in part by the Netherlands Organization for Scientific Research (NWO) under project numbers VI.C.192.080 and 2023.017; and is part of
‘Hybrid Intelligence: augmenting human intellect’
(https://hybrid-intelligence-centre.nl) with project number 024.004.022 of the
research program `Gravitation' which is (partly) financed by the Dutch Research
Council (NWO).

We kindly thank Maya K. Nachesa, Annika Kniele and Wafaa Mohammed for their contributions to evaluation. We thank Zena Al-Khalili and members of the LTL lab for their helpful feedback.

\bibliography{anthology1,anthology2,custom} 

\clearpage

\appendix
\label{sec:appendix}

\section{Prompts}\label{app:prompts}
\subsection{Prompt for parallel}
\label{app-1.1}

We provide the prompt used for parallel sampling here. Note that we explicitly instruct the model not to generate any additional content beyond the translation itself. In addition, we disable thinking mode by explicitly specifying this requirement in the system prompt. This prompt is shared with the first round of sequential sampling, ensuring strict comparability; see \Cref{app-1.2}.

\begin{greenprompt}
\begin{Verbatim}[
  breaklines=true,
  breakanywhere=true,
  breaksymbolright={},
  breaksymbolsepleft=0pt,
  breaksymbolindent=0pt,
  breaksymbolleft={},
  fontsize=\small
]

[system]
You are a professional translator. You will be asked to conduct translation-related tasks. Output ONLY the transaltion segment - do NOT generate additional content. Do not include any <think> tags. Do not show your reasoning.

[user turn initial]
Provide only one translation and do not output anything else after that.
Translate the following text from English to {target language}.

English: {source}

[assistant turn]
{assistant response}
\end{Verbatim}
\end{greenprompt}

\subsection{Prompt for multi-turn sequential sampling (improving instruction)}
\label{app-1.2}

We provide the multi-turn prompt used for sequential sampling here. Note that the prompt for the first round is identical to that in \Cref{app-1.1}, as shown in green. The iterative refinement instruction is shown in red, following \citet{wu2025please}, where we ask the model to ``translate again'' for a better version.

\begin{greenprompt}
\begin{Verbatim}[breaklines=true, 
                 breakanywhere=true,
                 breaksymbolright={},
                 breaksymbolsepleft=0pt,
                 breaksymbolindent=0pt,
                 breaksymbolleft={},
                 fontsize=\small]

[system]
You are a professional translator. You will be asked to conduct translation-related tasks. Output ONLY the transaltion segment - do NOT generate additional content. Do not include any <think> tags. Do not show your reasoning.

[user turn initial]
Provide only one translation and do not output anything else after that.
Translate the following text from English to {target language}.

English: {source}

[assistant turn]
{assistant response}
\end{Verbatim}
\end{greenprompt}
\vspace{1mm}
\begin{redprompt}
\begin{Verbatim}[
  breaklines=true,
  breakanywhere=true,
  breaksymbolright={},
  breaksymbolsepleft=0pt,
  breaksymbolindent=0pt,
  breaksymbolleft={},
  fontsize=\small
]

[user turn revisiting]
Please translate again for a better version.

[assistant turn]
{assistant response}

[user turn revisiting]
Please translate again for a better version.

...
\end{Verbatim}
\end{redprompt}

\section{Generation Details \label{app:generation_details}}
For the open-source models, we generate samples using \texttt{vLLM v0.11.1} \citep{kwon2023efficient} inference engine. We run \texttt{Qwen3-32B} on $4$ \texttt{NVIDIA RTX A6000} cards with up to $8$ request processing at the same time.  With the total budget $30$, each full-context sequential run, for a single language pair takes on average $14$h; each context-1 sequential sampling run takes slightly more that $4$h, and each parallel run takes on average $4$h.
We run \texttt{Qwen3-4B-Instruct-2507} on $2$ \texttt{NVIDIA RTX A6000} cards processing on average $40$ requests in parallel. Context-$5$ sequential run takes on average $1$h $18$m, and each parallel run takes on average $20$m.
Both models are run with the generation parameters, reported in \Cref{tab:model-params}.

Regarding the KV-cache friendliness, full-context sequential sampling can naturally reuse the previous cached activations, whereas context-$k$ sampling requires  cache recomputation, with context-$k$ requiring recomputing only one previous outputs. Parallel sampling enjoys reusing the KV-cache of the shared prefixes (\eg prompt).
\begin{table}[h]
\centering
\begin{tabular}{l r}
\hline
\textbf{Parameter} & \textbf{Value} \\
\hline
\textit{max-model-len} & $32768$ \\
\textit{max-num-batched-tokens} & $16392$ \\
\textit{max-num-seqs} & $8$ \\
\textit{max-tokens} & $4096$ \\
top-$k$ & $20$ \\
top-$p$ & $0.8$ \\
\hline
\end{tabular}
\caption{\texttt{Qwen3-*} generation parameters.}
\label{tab:model-params}
\end{table}

We run the \texttt{GPT-4o-mini} using official API from the model provider, using default generation parameters for this model (no top-$p$, no top-$k$ filtering, temperature $1.0$). For this model, we slightly edit the prompt, by replacing ``English: \{source\}" line simply with ``\{source\}", otherwise model sometimes includes target language name in its outputs.

We run the evaluation models on a single \texttt{NVIDIA RTX A6000} GPU.
We use batch size $8$ for COMET (\texttt{wmt22-comet-da}), COMET-KIWI (\texttt{wmt23-cometkiwi-da-xl}), and MetricX (\texttt{metricx-24-hybrid-xxl-v2p6-bfloat16}) metrics, and batch size $32$ for the ReMedy (\texttt{ReMedy-9B-22}) metrics. For MetricX, we set max input length as $1536$, and for ReMedy as $4096$. For ReMedy, we do not use callibration, as this feature is needed only to spread out the score distribution. Other metric parameters are as default.

For our main experiments, we compare parallel and sequential under a high-diversity/creativity regime using temperature $1.0$. Temperature sampling, while increasing diversity, is well known to incur more errors due to random sampling  \citep{troshin2025control}. We expect a good selector model to effectively ignore samples with translation errors. Empirically, we observe better results with lower temperature for sequential sampling without a selector (\Cref{fig:ablation:temp}), which highlights this phenomenon.

\section{Human Annotation \label{app:annotation}}
\begin{figure*}[t]
    \centering
    \setlength{\tabcolsep}{2pt}

     \includegraphics[width=1.0\textwidth]{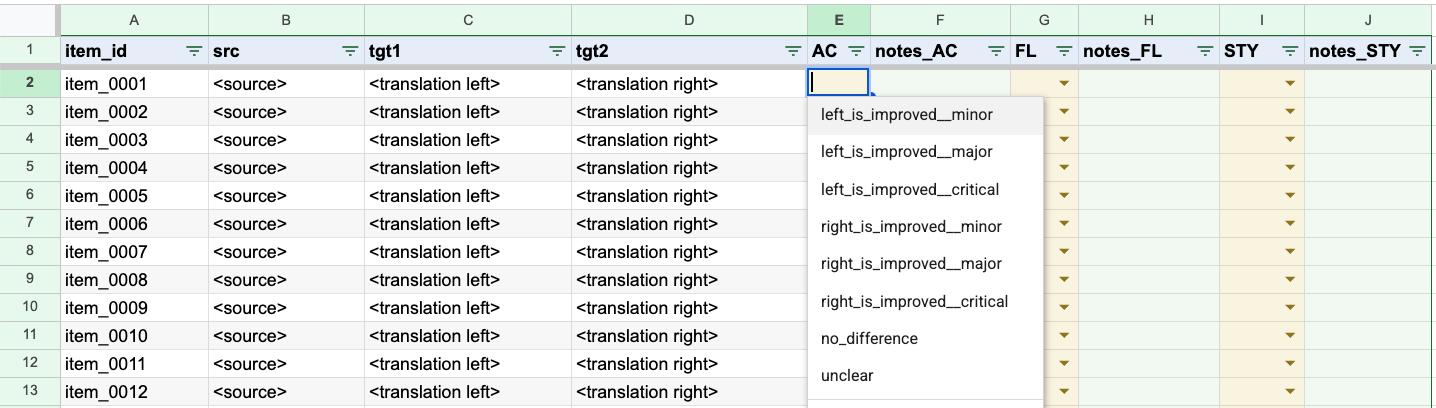}

    \caption{A demonstration of the human annotation interface using Google Sheets. For visibility of the demonstration, we replace the exact sentences with the placehoders for the source, left and right translations. Translations are presented in a shuffled way to mitigate the position bias for human annotators.}
    \label{fig:annotation_interface}
\end{figure*}

Annotators are native speakers of a target language (researchers, students, university workers), and participate on a voluntary, unpaid basis. They were provided with the labeling instructions(\Cref{app:annotation}, and then asked to annotate which one of the two translations is better for each quality dimension: accuracy, fluency, and style, given the source sentence in English. For our pilot human annotation experiments, we use a blind comparison setup, \ie, annotators see a shuffled pairs of target segments. We randomly select $100$ instances from the WMT24++ segment-level dataset for each annotation experiment. Source sentence ids are the same across target languages for each experiment type, but the source sentences do not intersect between different experiments to prevent memorization biases and model guessing. We use a simple tabular interface via Google Sheets as a tool for collecting labels, where annotators select the labels using a predefined set of labels defined in \Cref{app:annotation}, and can optionally leave the notes, explaining their decisions. We include a screenshot of our interface in \Cref{fig:annotation_interface}.
We present the human annotation instructions in \Cref{fig:human_annotation_instructions}.

\begin{figure*}[t]
\begin{framed}
\vspace{-3mm}
\begin{Verbatim}[
    breaklines=true,
    breakanywhere=true,
    breaksymbolright={},
    breaksymbolsepleft=0pt,
    breaksymbolindent=0pt,
    breaksymbolleft={},
    fontsize=\smaller
]

Use the source text as the reference and compare translation1 and translation2 for each MQM dimension separately: AC accuracy, FL fluency, and STY style. Assign one label per dimension.

Choose exactly one label:

left_is_improved_minor
left_is_improved_major
left_is_improved_critical
right_is_improved_minor
right_is_improved_major
right_is_improved_critical
no_difference
unclear

Here, left = translation1 and right = translation2.

Use left_is_improved if translation1 is better.
Use right_is_improved if translation2 is better.
Use no_difference if both translations are equivalent for that dimension, including when both are equally good or equally bad.
Use unclear only when the source or translations do not provide enough information to make a reliable judgment, or both translations have positive and negative aspects.

Severity:
minor = small improvement that does not substantially affect meaning, fluency, or appropriateness.
major = clear improvement that noticeably affects meaning, fluency, or style.
critical = improvement fixes or avoids a severe issue that makes the translation seriously misleading, unusable, or inappropriate.

AC — Accuracy: judge whether the translation correctly preserves the source meaning. Consider distortions, omissions, and additions.
FL — Fluency: judge linguistic well-formedness in the target language. Consider grammar, syntax, spelling, punctuation, and mechanical correctness.
STY — Style: judge whether the language style/use is appropriate for the context, even if grammatically correct.

\end{Verbatim}
\vspace{-3mm}
\end{framed}
\caption{Human annotation instructions. \label{fig:human_annotation_instructions}}
\end{figure*}

\section{Experimental Details for \Cref{sec:6-2}}
\label{app:D}
We conduct our experiments on \texttt{Qwen-3-32B} model. The self-refinement setting is identical to that in \Cref{sec:5}.

The WMT24++ dataset consists of 171 documents and 998 sentences (some of which are short paragraphs). We use the metadata\footnote{https://huggingface.co/datasets/google/wmt24pp} provided by WMT24++ to preserve the sentence-to-document mapping. Based on this mapping, we construct 633 segments (two consecutive sentences) following a sliding window setting, as described in \Cref{sec:5}. For each document, we remove one HEAD and one TAIL sentence to ensure strict comparability.

Note that we use a special separator "\textbackslash t" to connect HEAD and TAIL sentences in each segment and split them when calculating the corresponding sentence-level score.

\section{Diversity Results \label{app:diversity}}

\begin{figure*}[t]
    \centering
    \setlength{\tabcolsep}{2pt}

     \includegraphics[width=0.6\textwidth]{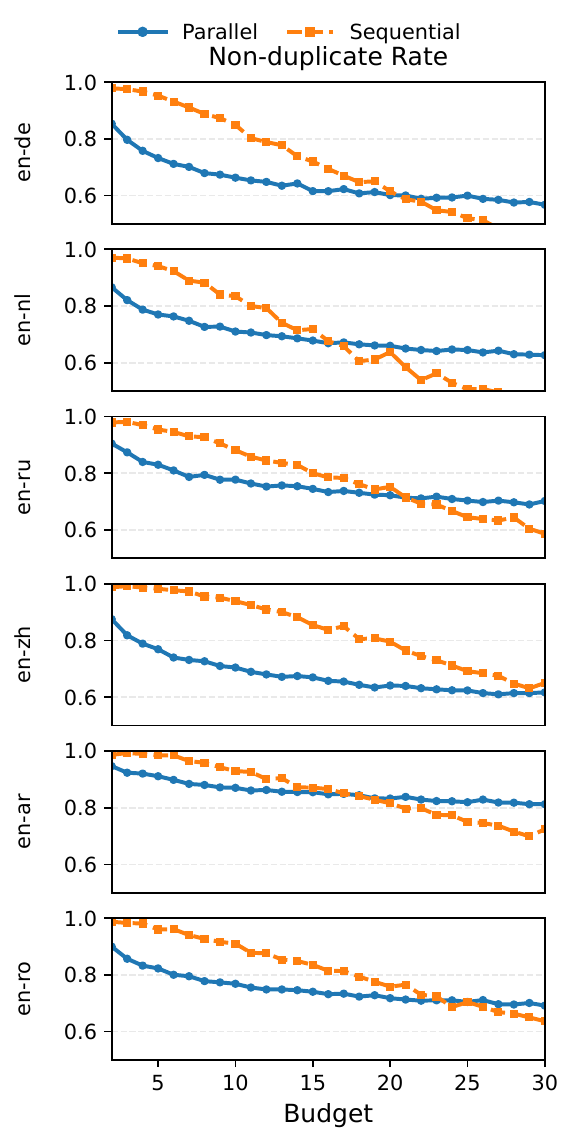}

    \caption{Non-duplicate rate for \texttt{Qwen-3-32B} model for parallel and sequential sampling per language. Diversity metric represents how close a new sample is to every previous sample using exact match similarity.}
    \label{fig:diversity_plots_per_language}
\end{figure*}

In \Cref{fig:diversity_plots_per_language}, we show the non-duplicate rate (marginal diversity) per language direction. Results demonstrate the uniformity of the diversity findings over the language directions: sequential produce less duplicates compared to parallel for earlier rounds of generation.

\section{Results}
\paragraph{Qwen3-32B results.}
\Cref{fig:main_plots_32b_model} presents main results across 6 language pairs for the \texttt{Qwen3-32B} model.

\Cref{tab:main_results} presents a compact view to easier compare different strategies 1) first steps of self-refinement, 2) sampling with sequential selector results, including the context window size ablation ($h$ = 'full', $h=5$, $h=1$).

\begin{figure*}[t]
    \centering
    \setlength{\tabcolsep}{2pt}

     \includegraphics[width=1.0\textwidth]{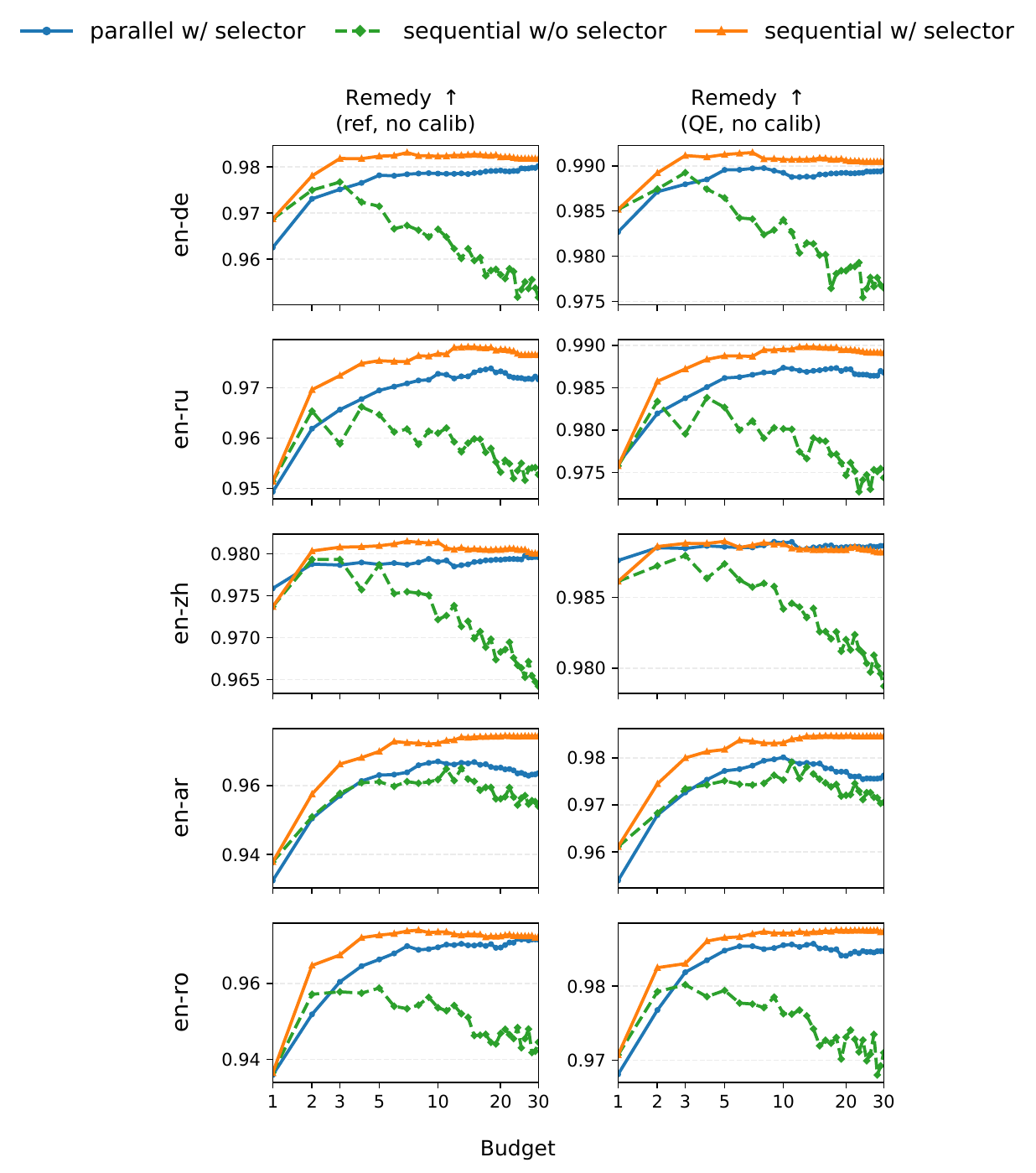}

    \caption{Additional evaluation with the ReMedy (\texttt{ReMedy-9B-22}) metric  \citep{tan-monz-2025-remedy} for 5 language directions (en-nl is not supported by the metric). Samples are from the \texttt{Qwen-3-32B} model. Quality vs.\ budget. Selector is \texttt{COMET-KIWI}.}
    \label{fig:main_results_with_Remedy}
\end{figure*}

\begin{table*}[t]
\centering

\begin{tabular}{@{}llrrr@{\hspace{1.2em}}rrr@{\hspace{1.2em}}>{\columncolor[RGB]{255,255,255}[\tabcolsep][0pt]}c@{}}
\toprule
 &  & \multicolumn{3}{c}{Seq.\ w/o\ selector} & \multicolumn{3}{c}{Seq.\ w/\ selector} & \multicolumn{1}{c}{Parallel} \\
Lang  & Metric          &                             $t=1$ &                             $t=2$ &                             $t=3$ &                        $h=\infty$ &                                     $h=5$ &                                      $h=1$ &                                            \\
\midrule
en-de & COMET           & \cellcolor[RGB]{255,255,255}81.26 & \cellcolor[RGB]{248,252,247}81.59 & \cellcolor[RGB]{250,253,249}81.49 & \cellcolor[RGB]{226,244,221}82.62 &         \cellcolor[RGB]{234,247,231}82.24 &          \cellcolor[RGB]{221,242,216}82.84 & \cellcolor[RGB]{217,240,211}\textbf{83.04} \\
      & metricx24 (ref) &  \cellcolor[RGB]{255,255,255}2.23 &  \cellcolor[RGB]{240,249,237}1.92 &  \cellcolor[RGB]{234,247,231}1.81 &  \cellcolor[RGB]{222,242,217}1.55 &          \cellcolor[RGB]{220,241,215}1.52 &  \cellcolor[RGB]{217,240,211}\textbf{1.46} &           \cellcolor[RGB]{229,245,225}1.70 \\
      & metricx24 (qe)  &  \cellcolor[RGB]{255,255,255}2.02 &  \cellcolor[RGB]{239,248,236}1.72 &  \cellcolor[RGB]{233,246,230}1.62 &  \cellcolor[RGB]{220,241,215}1.38 &          \cellcolor[RGB]{219,241,213}1.36 &  \cellcolor[RGB]{217,240,211}\textbf{1.32} &           \cellcolor[RGB]{228,244,224}1.53 \\
\midrule
en-nl & COMET           & \cellcolor[RGB]{255,255,255}82.10 & \cellcolor[RGB]{244,251,242}82.81 & \cellcolor[RGB]{246,251,244}82.70 & \cellcolor[RGB]{226,244,222}83.92 &         \cellcolor[RGB]{225,243,220}84.02 & \cellcolor[RGB]{217,240,211}\textbf{84.52} &          \cellcolor[RGB]{220,241,215}84.30 \\
      & metricx24 (ref) &  \cellcolor[RGB]{255,255,255}3.94 &  \cellcolor[RGB]{240,249,237}3.33 &  \cellcolor[RGB]{235,247,232}3.16 &  \cellcolor[RGB]{221,241,215}2.57 &          \cellcolor[RGB]{218,240,212}2.47 &  \cellcolor[RGB]{217,240,211}\textbf{2.42} &           \cellcolor[RGB]{228,244,224}2.87 \\
      & metricx24 (qe)  &  \cellcolor[RGB]{255,255,255}3.63 &  \cellcolor[RGB]{239,249,237}3.00 &  \cellcolor[RGB]{235,247,232}2.85 &  \cellcolor[RGB]{221,241,215}2.26 &          \cellcolor[RGB]{218,241,213}2.18 &  \cellcolor[RGB]{217,240,211}\textbf{2.12} &           \cellcolor[RGB]{229,245,225}2.60 \\
\midrule
en-ru & COMET           & \cellcolor[RGB]{255,255,255}80.98 & \cellcolor[RGB]{237,248,234}82.44 & \cellcolor[RGB]{238,248,235}82.35 & \cellcolor[RGB]{221,241,215}83.70 &         \cellcolor[RGB]{223,242,218}83.51 & \cellcolor[RGB]{217,240,211}\textbf{84.00} &          \cellcolor[RGB]{227,244,222}83.21 \\
      & metricx24 (ref) &  \cellcolor[RGB]{255,255,255}4.41 &  \cellcolor[RGB]{237,248,234}3.60 &  \cellcolor[RGB]{233,246,230}3.45 &  \cellcolor[RGB]{220,241,215}2.88 &          \cellcolor[RGB]{219,241,213}2.82 &  \cellcolor[RGB]{217,240,211}\textbf{2.73} &           \cellcolor[RGB]{230,245,226}3.29 \\
      & metricx24 (qe)  &  \cellcolor[RGB]{255,255,255}3.86 &  \cellcolor[RGB]{237,248,234}3.09 &  \cellcolor[RGB]{234,247,231}2.96 &  \cellcolor[RGB]{221,242,216}2.41 &          \cellcolor[RGB]{219,241,213}2.31 &  \cellcolor[RGB]{217,240,211}\textbf{2.22} &           \cellcolor[RGB]{229,245,225}2.76 \\
\midrule
en-zh & COMET           & \cellcolor[RGB]{255,255,255}84.85 & \cellcolor[RGB]{246,251,244}85.14 & \cellcolor[RGB]{248,252,247}85.08 & \cellcolor[RGB]{238,248,235}85.40 &         \cellcolor[RGB]{232,246,228}85.59 &          \cellcolor[RGB]{225,243,221}85.79 & \cellcolor[RGB]{217,240,211}\textbf{86.06} \\
      & metricx24 (ref) &  \cellcolor[RGB]{255,255,255}2.47 &  \cellcolor[RGB]{236,247,233}2.20 &  \cellcolor[RGB]{237,248,234}2.21 &  \cellcolor[RGB]{223,243,219}2.02 &          \cellcolor[RGB]{221,242,215}1.98 &  \cellcolor[RGB]{217,240,211}\textbf{1.92} &           \cellcolor[RGB]{232,246,228}2.13 \\
      & metricx24 (qe)  &  \cellcolor[RGB]{255,255,255}2.39 &  \cellcolor[RGB]{237,248,234}2.14 &  \cellcolor[RGB]{237,248,234}2.14 &  \cellcolor[RGB]{222,242,217}1.93 &          \cellcolor[RGB]{220,241,215}1.90 &  \cellcolor[RGB]{217,240,211}\textbf{1.85} &           \cellcolor[RGB]{232,246,229}2.07 \\
\midrule
en-ar & COMET           & \cellcolor[RGB]{255,255,255}76.11 & \cellcolor[RGB]{242,250,240}77.30 & \cellcolor[RGB]{240,249,237}77.54 & \cellcolor[RGB]{220,241,215}79.37 &         \cellcolor[RGB]{220,241,215}79.40 & \cellcolor[RGB]{217,240,211}\textbf{79.69} &          \cellcolor[RGB]{226,244,221}78.84 \\
      & metricx24 (ref) &  \cellcolor[RGB]{255,255,255}4.92 &  \cellcolor[RGB]{242,250,240}4.25 &  \cellcolor[RGB]{237,248,234}3.99 &  \cellcolor[RGB]{219,241,213}3.10 & \cellcolor[RGB]{217,240,211}\textbf{3.01} &           \cellcolor[RGB]{217,240,211}3.03 &           \cellcolor[RGB]{229,245,225}3.64 \\
      & metricx24 (qe)  &  \cellcolor[RGB]{255,255,255}4.37 &  \cellcolor[RGB]{242,250,240}3.72 &  \cellcolor[RGB]{236,248,233}3.44 &  \cellcolor[RGB]{218,241,213}2.56 & \cellcolor[RGB]{217,240,211}\textbf{2.49} &           \cellcolor[RGB]{218,240,212}2.52 &           \cellcolor[RGB]{229,245,225}3.08 \\
\midrule
en-ro & COMET           & \cellcolor[RGB]{255,255,255}82.68 & \cellcolor[RGB]{245,251,243}83.41 & \cellcolor[RGB]{242,250,240}83.60 & \cellcolor[RGB]{228,244,224}84.65 &         \cellcolor[RGB]{225,243,221}84.84 & \cellcolor[RGB]{217,240,211}\textbf{85.46} &          \cellcolor[RGB]{219,241,213}85.34 \\
      & metricx24 (ref) &  \cellcolor[RGB]{255,255,255}5.90 &  \cellcolor[RGB]{241,249,238}5.10 &  \cellcolor[RGB]{237,248,234}4.88 &  \cellcolor[RGB]{221,241,215}3.99 &          \cellcolor[RGB]{218,241,213}3.87 &  \cellcolor[RGB]{217,240,211}\textbf{3.80} &           \cellcolor[RGB]{229,245,225}4.44 \\
      & metricx24 (qe)  &  \cellcolor[RGB]{255,255,255}5.39 &  \cellcolor[RGB]{242,250,239}4.65 &  \cellcolor[RGB]{236,248,234}4.37 &  \cellcolor[RGB]{220,241,215}3.47 &          \cellcolor[RGB]{219,241,213}3.40 &  \cellcolor[RGB]{217,240,211}\textbf{3.30} &           \cellcolor[RGB]{230,245,226}4.02 \\
\bottomrule
\end{tabular}

\caption{Automatic evaluation results for sequential and parallel sampling strategies across 6 languages for the \texttt{Qwen3-32B} model. Left side presents results for turn-$1,2,3,$ of sequential sampling, right side shows results with Best-of-$N$ sampling with budget $30$ for both sequential (full context, $h=5$, $h=1$) and parallel with \texttt{COMET-KIWI} selector. The best results in each row are marked with \textbf{bold}. \label{tab:main_results_full}}
\end{table*}

\begin{figure*}[t]
    \centering
    \setlength{\tabcolsep}{2pt}

     \includegraphics[width=1.0\textwidth]{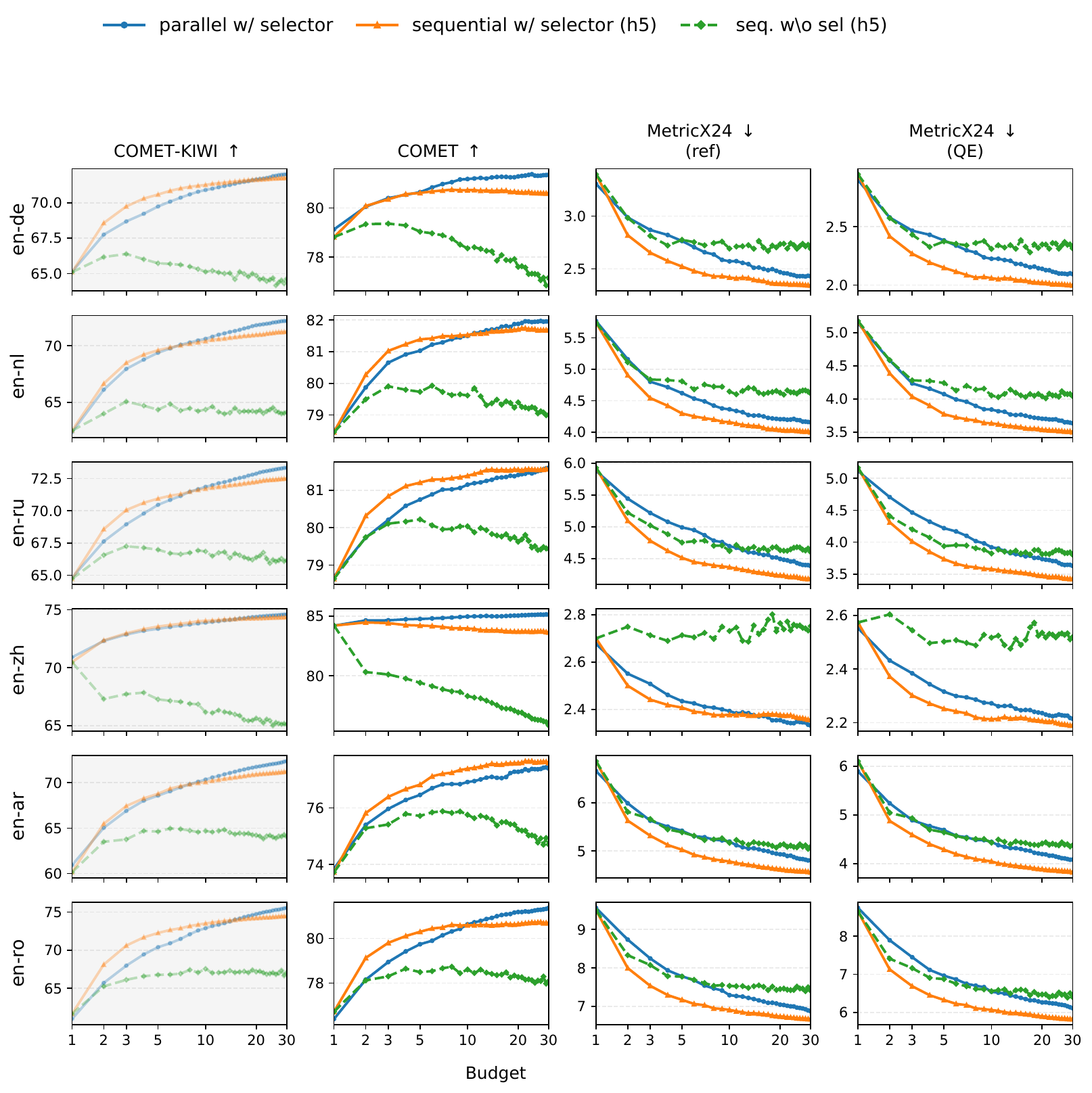}

    \caption{Quality vs budget results for the \texttt{Qwen-3-4B-2507} model. Selector is \texttt{COMET-KIWI}.}
    \label{fig:main_plots_4b_model}
\end{figure*}

\paragraph{Qwen3-4b-instruct additional results.}
Results for the \texttt{Qwen3-4b-instruct} are presented in  \Cref{fig:main_plots_4b_model}.

\begin{figure*}[t]
    \centering
    \setlength{\tabcolsep}{2pt}

     \includegraphics[width=1.0\textwidth]{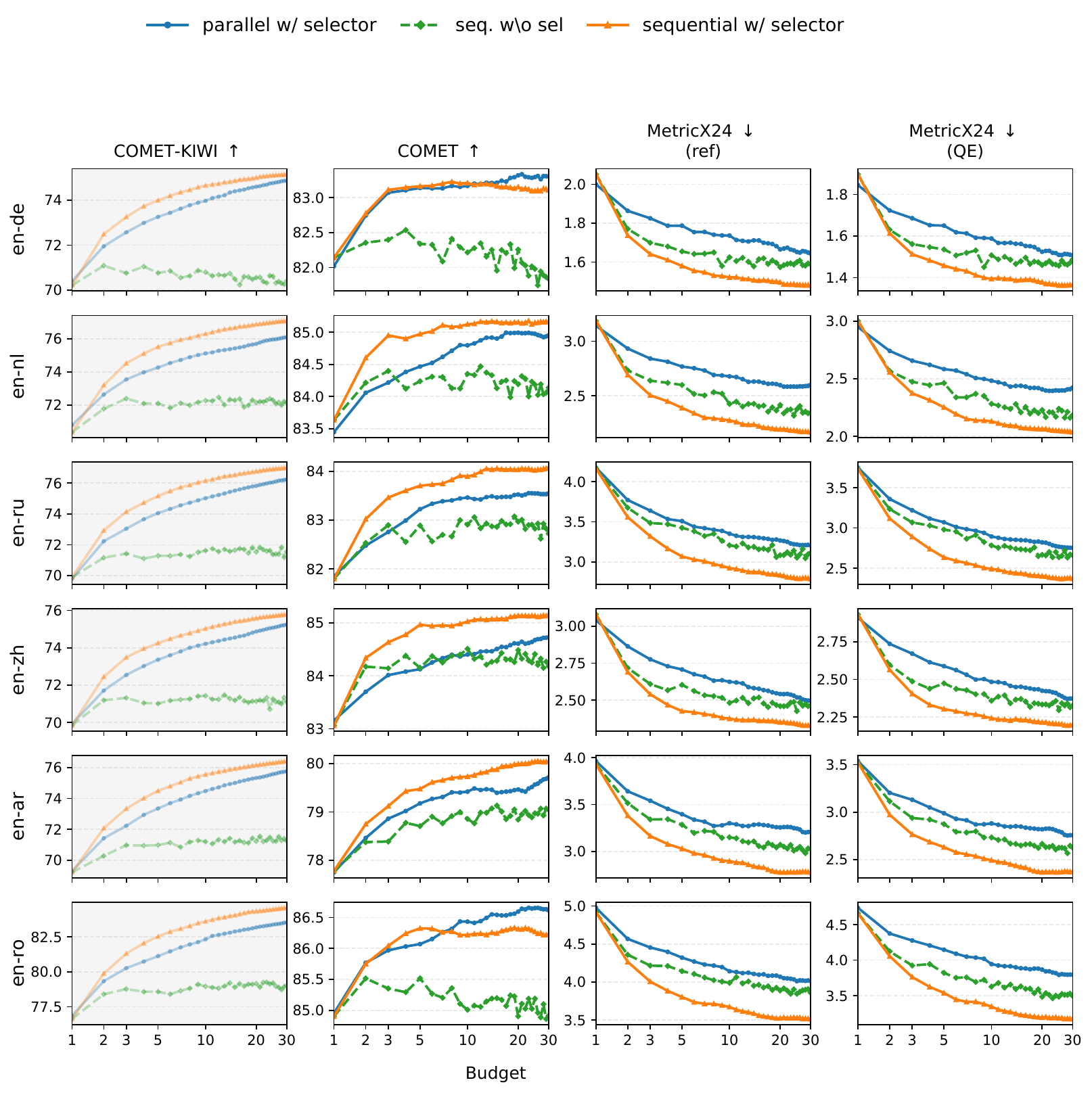}

    \caption{Quality vs budget results for the \texttt{GPT-4o-mini} model. Selector is \texttt{COMET-KIWI}.}
    \label{fig:gpt_4o_mini}
\end{figure*}

\begin{figure*}[t]
    \centering
    \setlength{\tabcolsep}{2pt}

     \includegraphics[width=1.0\textwidth]{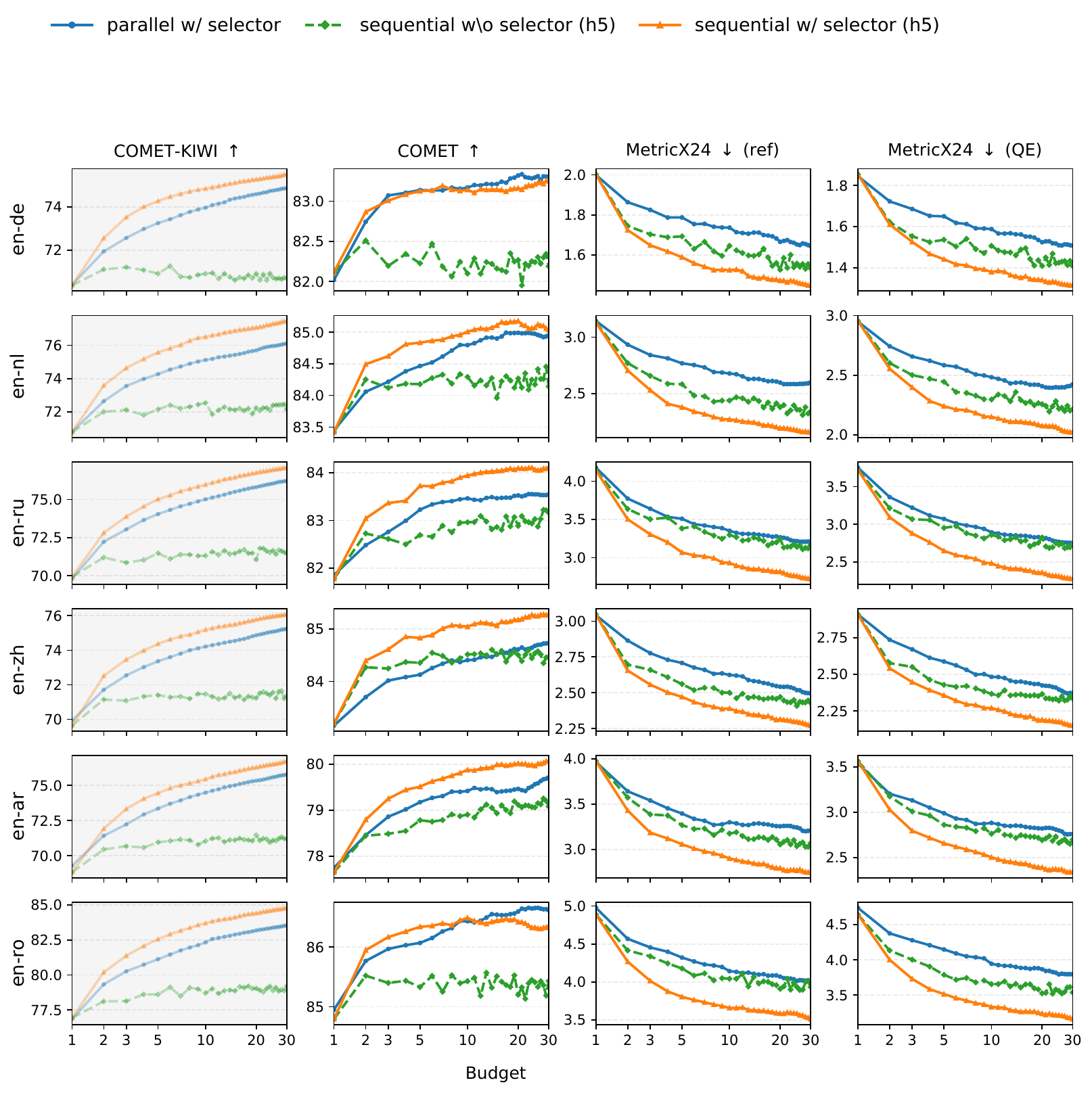}

    \caption{Quality vs budget results for the \texttt{GPT-4o-mini} model. Sequential sampling with context-5 ablation. Selector is \texttt{COMET-KIWI}.}
    \label{fig:gpt_4o_mini_history_5}
\end{figure*}

\begin{figure*}[t]
    \centering
    \setlength{\tabcolsep}{2pt}

     \includegraphics[width=1.0\textwidth]{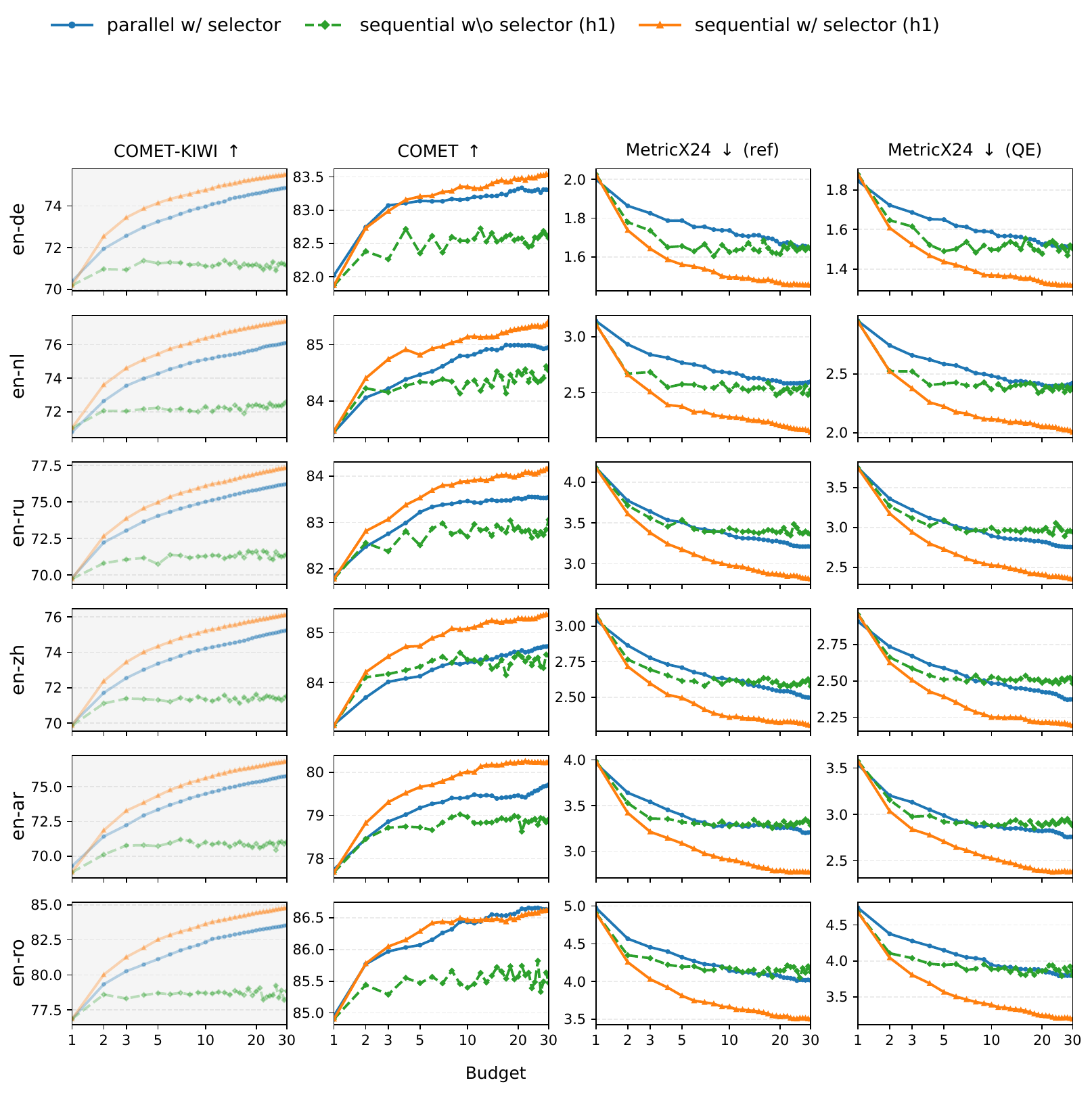}

    \caption{Quality vs budget results for the \texttt{GPT-4o-mini} model. Sequential sampling with context-1 ablation. Selector is \texttt{COMET-KIWI}.}
    \label{fig:gpt_4o_mini_history_1}
\end{figure*}

\paragraph{GPT-4o-mini additional results.}
Results with full-context sequential sampling for the \texttt{GPT-4o-mini} are presented in \Cref{fig:gpt_4o_mini}. The context-5 ablation is presented in \Cref{fig:gpt_4o_mini_history_5}, and the context-1 ablation is presented in \Cref{fig:gpt_4o_mini_history_1}.

\section{Human Evaluation Results \label{app_human_annotation_results}}
For the human evaluation, we present the full results with a split by language directions in \Cref{fig:human_annotation_plot_full}. When comparing sequential w\ selector, or sequential turn-2 vs turn-1, we notice, that sequential sampling improves fluency, accuracy and style; but the quality improvements can vary across languages and quality dimensions. When comparing sequential vs parallel under budget 30, we notice that parallel often demonstrates higher accuracy. 

\begin{figure*}[th]
\centering
\smaller

\includegraphics[width=0.7\textwidth]{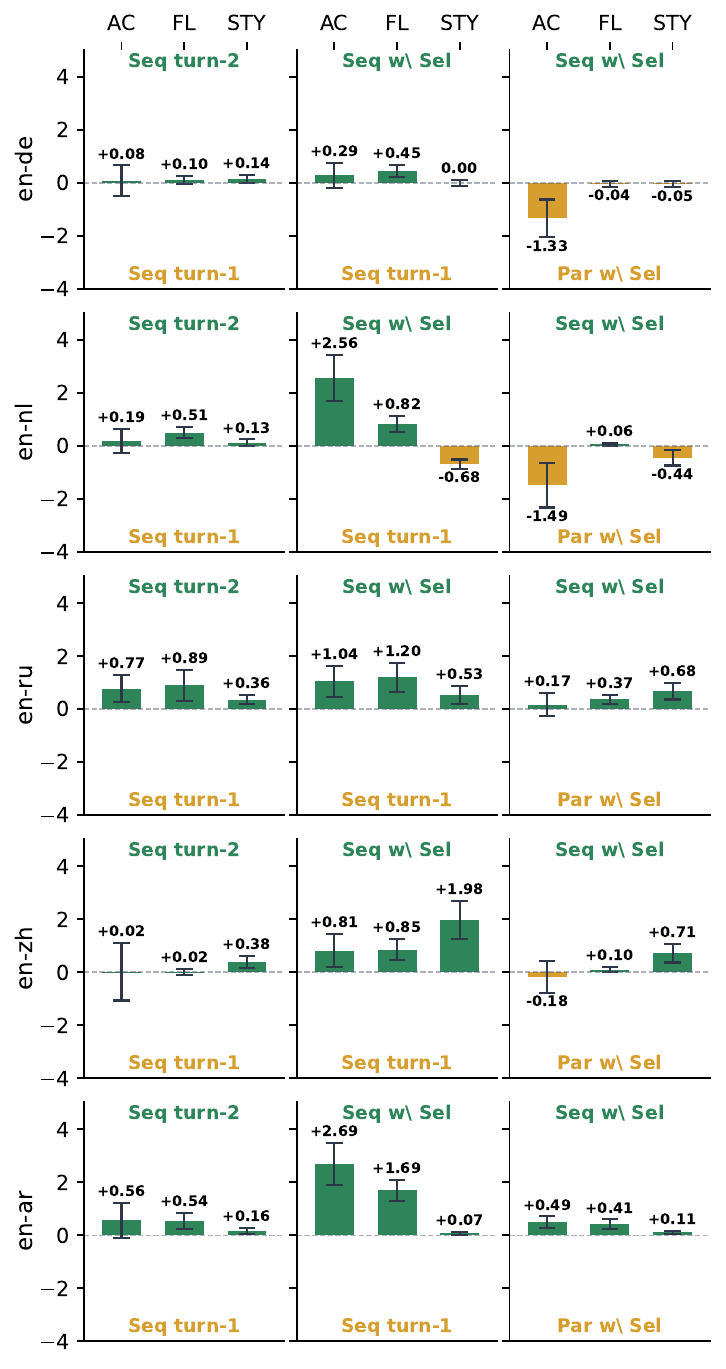}

\caption{Human annotation results using MQM \textbf{average weighted relative improvement} for accuracy (AC), fluency (FL) and style (STY) for the for the \texttt{Qwen3-32B} model. Results are presented for (i) turn-$1$ vs turn-$2$, (ii) Sequential w\ Sel (budget 30) vs Seq. turn-1, (iii)  Sequential w\ Sel (budget 30) vs Parallel w\ Sel (budget 30). Error bars represent standard error of the mean. \label{fig:human_annotation_plot_full-2}}

\end{figure*}

\begin{table*}[t]
\centering

\begin{tabular}{ll|rrrr|rrrr|rrrr}
\hline
      &          & \multicolumn{4}{c|}{turn1 vs turn2} & \multicolumn{4}{c|}{turn1 vs seq.\ w/\ sel} & \multicolumn{4}{c}{par.\ w/\ sel vs seq.\ w/\ sel} \\
Lang. & Metric   &  minr &  majr & crit & nd &  minr & majr & crit & nd &  minr & majr & crit & nd \\
\hline
en-de & Accuracy &  10/5 &   3/5 &  1/1 & 41 &   7/3 &  2/1 &  0/1 & 42 &  10/6 &  4/1 &  2/0 & 29 \\
      & Fluency  &   9/6 &   1/3 &  0/0 & 51 &   4/5 &  1/6 &  0/0 & 42 &   3/6 &  1/0 &  0/0 & 47 \\
      & Style    &  7/17 &   2/2 &  0/0 & 41 &  5/10 &  1/0 &  0/0 & 43 &  9/11 &  1/0 &  0/0 & 36 \\
\hline
en-nl & Accuracy & 20/19 & 11/15 &  1/1 & 31 & 10/20 & 3/12 &  1/7 & 27 & 20/17 &  8/6 &  5/1 & 19 \\
      & Fluency  & 11/25 &  4/11 &  0/0 & 46 &  9/22 &  1/9 &  0/1 & 53 &  6/12 &  0/0 &  0/0 & 78 \\
      & Style    &   5/3 &   1/4 &  0/0 & 86 & 32/11 & 10/1 &  0/0 & 43 & 29/21 &  4/2 &  1/0 & 40 \\
\hline
en-ru & Accuracy &  12/6 &   5/5 &  0/3 & 59 &  9/11 &  8/7 &  0/4 & 54 &  18/4 & 6/12 &  1/1 & 51 \\
      & Fluency  &   3/4 &   3/9 &  1/3 & 68 &   0/8 & 3/15 &  1/3 & 68 &   0/6 &  2/8 &  0/0 & 82 \\
      & Style    & 10/19 &   2/7 &  0/0 & 56 & 16/17 & 2/17 &  1/0 & 43 & 15/17 & 3/11 &  0/1 & 52 \\
\hline
en-zh & Accuracy &   8/4 &   6/2 &  3/4 & 36 & 16/14 & 9/14 &  1/3 & 33 & 21/10 &  8/7 &  2/2 & 41 \\
      & Fluency  &   1/2 &   1/1 &  0/0 & 57 &  2/10 &  1/5 &  0/2 & 72 &   1/5 &  1/2 &  0/0 & 81 \\
      & Style    &   5/9 &   2/6 &  0/0 & 41 &  2/23 & 6/14 &  1/6 & 42 &  7/16 & 4/10 &  0/1 & 52 \\
\hline
en-ar & Accuracy &  6/11 &  6/16 &  3/3 & 53 &  4/10 & 1/13 &  1/9 & 61 &  10/7 & 4/14 &  0/0 & 61 \\
      & Fluency  &  5/17 &   3/6 &  0/1 & 65 &  5/16 & 1/22 &  0/2 & 52 &  5/11 & 3/10 &  0/0 & 70 \\
      & Style    &   5/6 &   1/4 &  0/0 & 81 &   0/2 &  0/1 &  0/0 & 96 &   1/7 &  0/1 &  0/0 & 90 \\
\hline
\end{tabular}

\caption{Raw MQM annotation counts for the directional minor improvements (minr), major improvements (majr), critical improvements (crit), and no difference (nd); */* formatting corresponds to the experiment  names in the first row of the figure; */* formatting for the relative scores means <n times left is improved>/<n times right is improved>. Annotation was only partially finished for en-de due to time constraints.}
\label{tab:mqm}
\end{table*}

\begin{figure*}[t]
    \centering
    \setlength{\tabcolsep}{2pt}

     \includegraphics[width=0.95\textwidth]{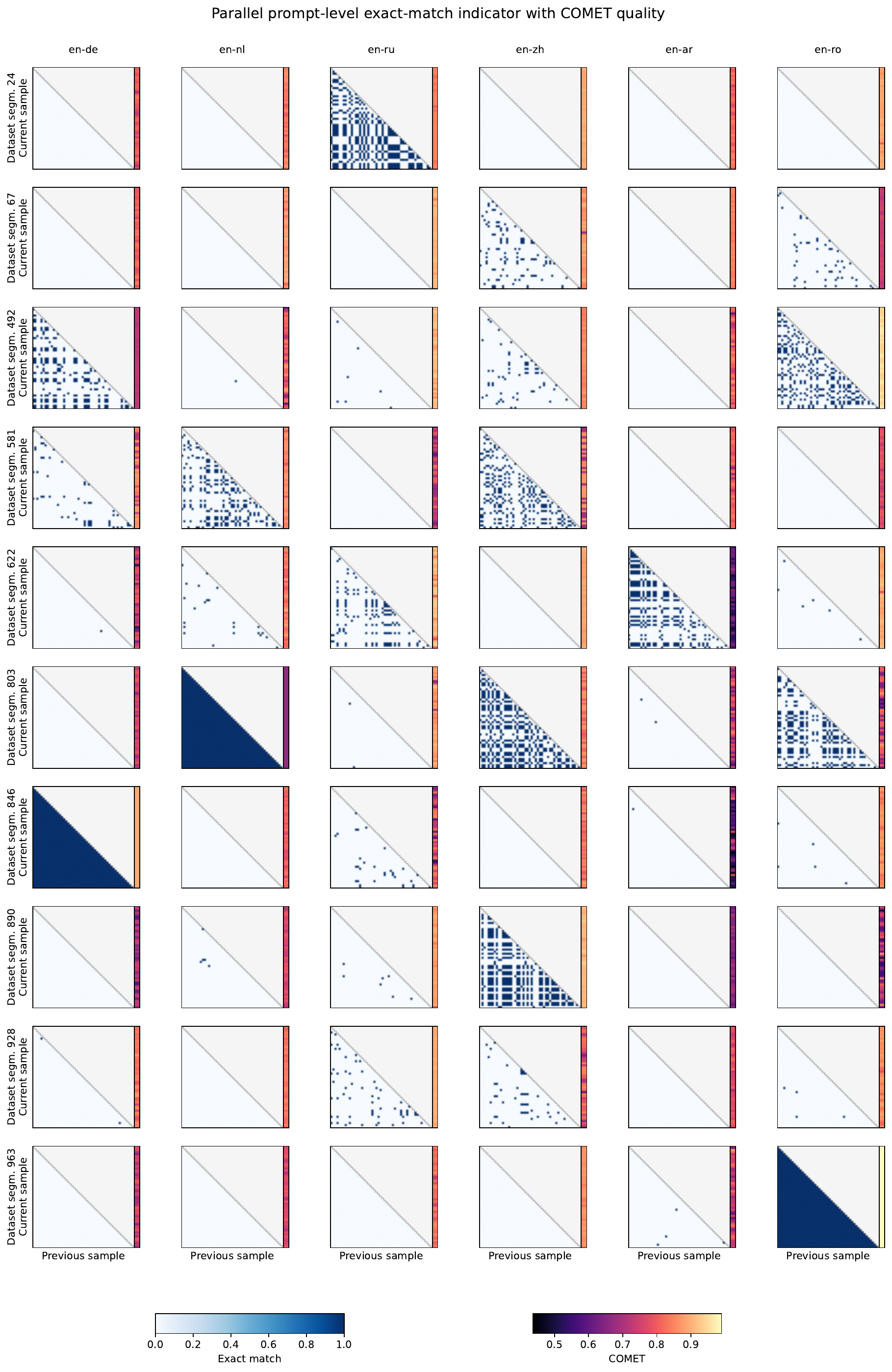}

    \caption{Diversity and MT quality (COMET) for \texttt{Qwen-3-32B} model for parallel sampling. Diversity metric represents how close a new sample is to every previous sample using exact match similarity.}
    \label{fig:diversity_plots_32b_model_EM_random_prompts_parallel}
\end{figure*}

\begin{figure*}[t]
    \centering
    \setlength{\tabcolsep}{2pt}

     \includegraphics[width=0.95\textwidth]{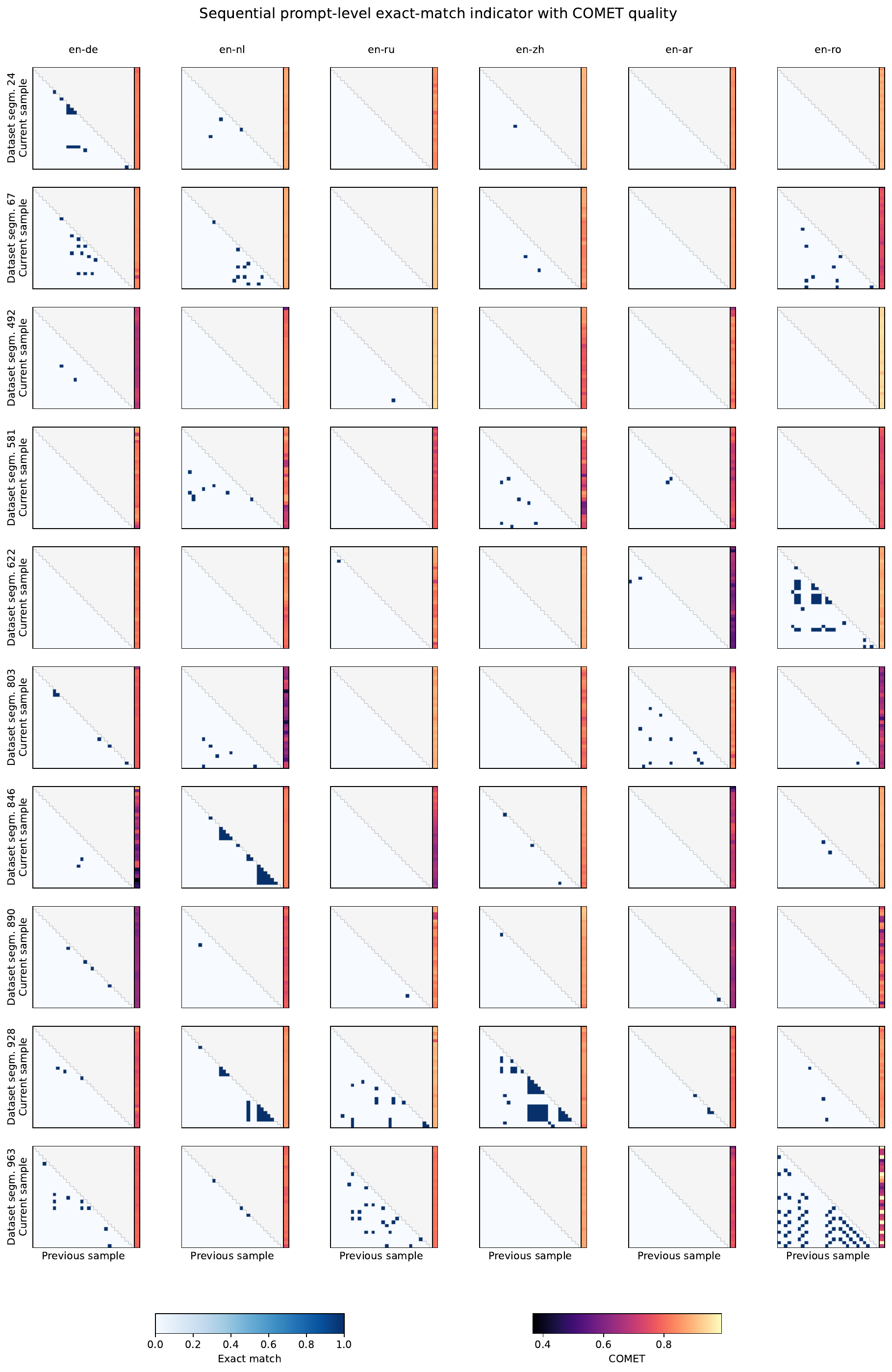}

    \caption{Diversity and MT quality (COMET) for \texttt{Qwen-3-32B} model for sequential sampling. Diversity metric represents how close a new sample is to every previous sample using exact match similarity.}
    \label{fig:diversity_plots_32b_model_EM_random_prompts_sequential}
\end{figure*}

\section{Extended Ablation Results}
\subsection{Context size ablation}
\begin{figure*}[t]
    \centering
    \setlength{\tabcolsep}{2pt}

     \includegraphics[width=1.0\textwidth]{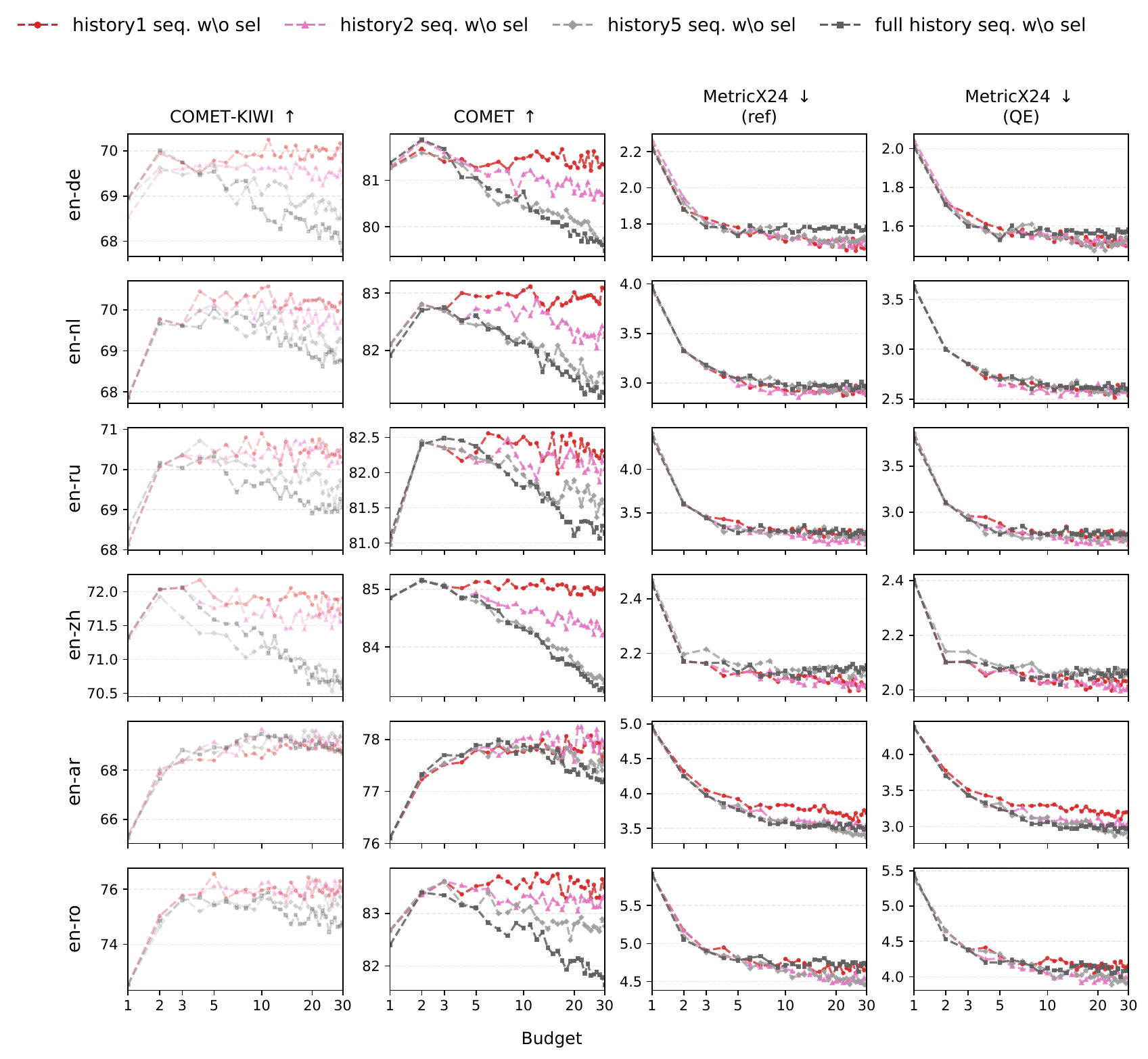}

    \caption{Context size ablation for the sliding window sequential sampling approach for the \texttt{Qwen-3-32B} model. The scaling of self-refinement (\ie \textbf{sequential sampling without selector}) can be largely improved by compacting the context using the sliding window approach.}
    \label{fig:ablation:history_ablation_without_sel}
\end{figure*}

\begin{figure*}[t]
    \centering
    \setlength{\tabcolsep}{2pt}

     \includegraphics[width=1.0\textwidth]{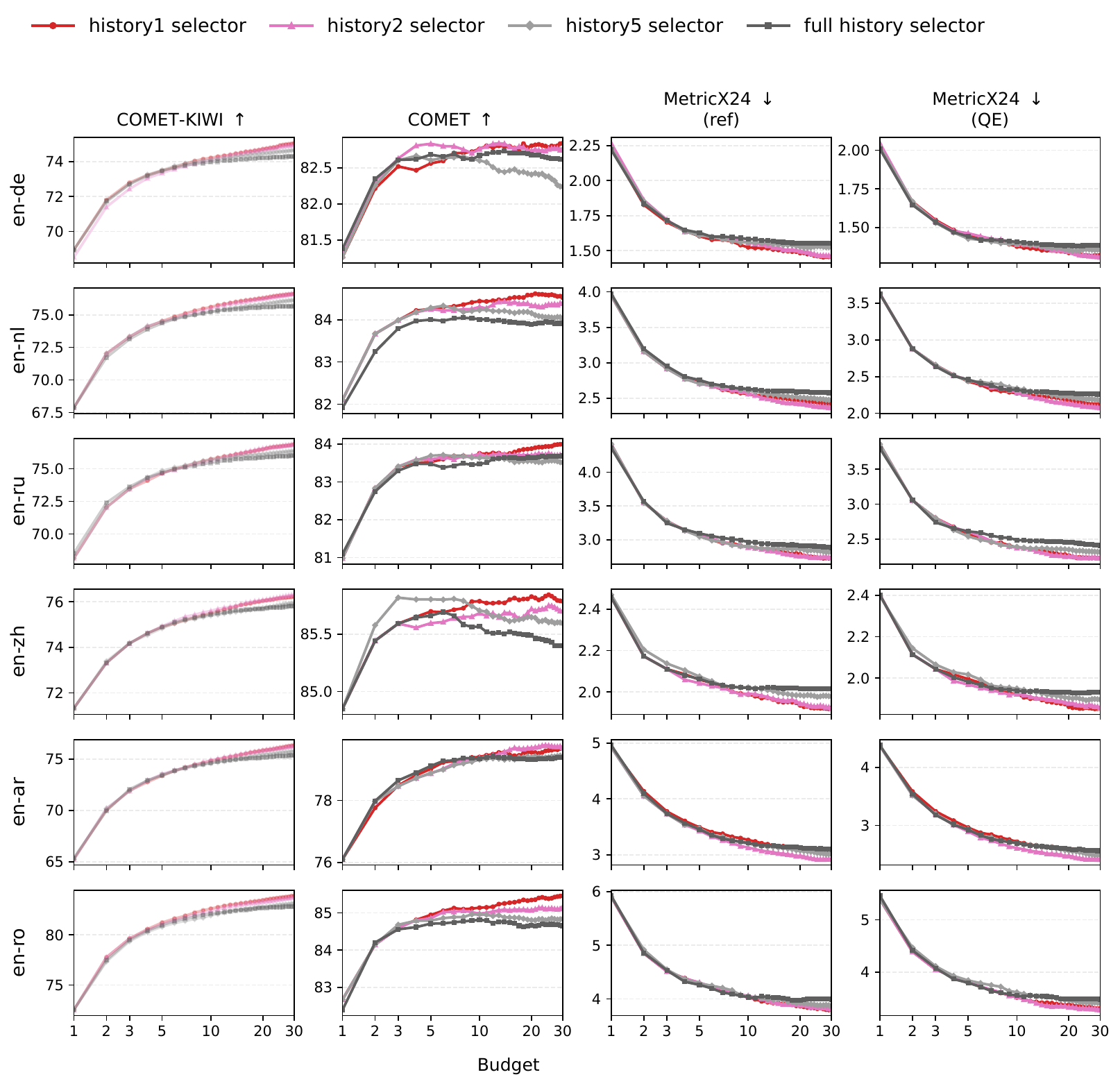}

    \caption{Context size ablation for the sliding window sequential sampling approach for the \texttt{Qwen-3-32B} model. The quality of \textbf{sequential sampling with selector} can be slightly improved by compacting the context using the sliding window approach.}
    \label{fig:ablation:history_ablation_with_sel}
\end{figure*}

In this section, we present the results of context managing for sequential sampling using the sliding window approach with context sizes: full context, context-5, context-2, context-1. Results without selector show substantial quality improvements in \Cref{fig:ablation:history_ablation_without_sel}, when constraining the context. \Cref{fig:ablation:history_ablation_with_sel} presents results for sequential sampling with selector, where we observe roughly the same pattern, but with a smaller difference between runs, smoothed by a selector. 

\subsection{Results with another selector metric}
\begin{figure*}[t]
    \centering
    \setlength{\tabcolsep}{2pt}

     \includegraphics[width=0.97\textwidth]{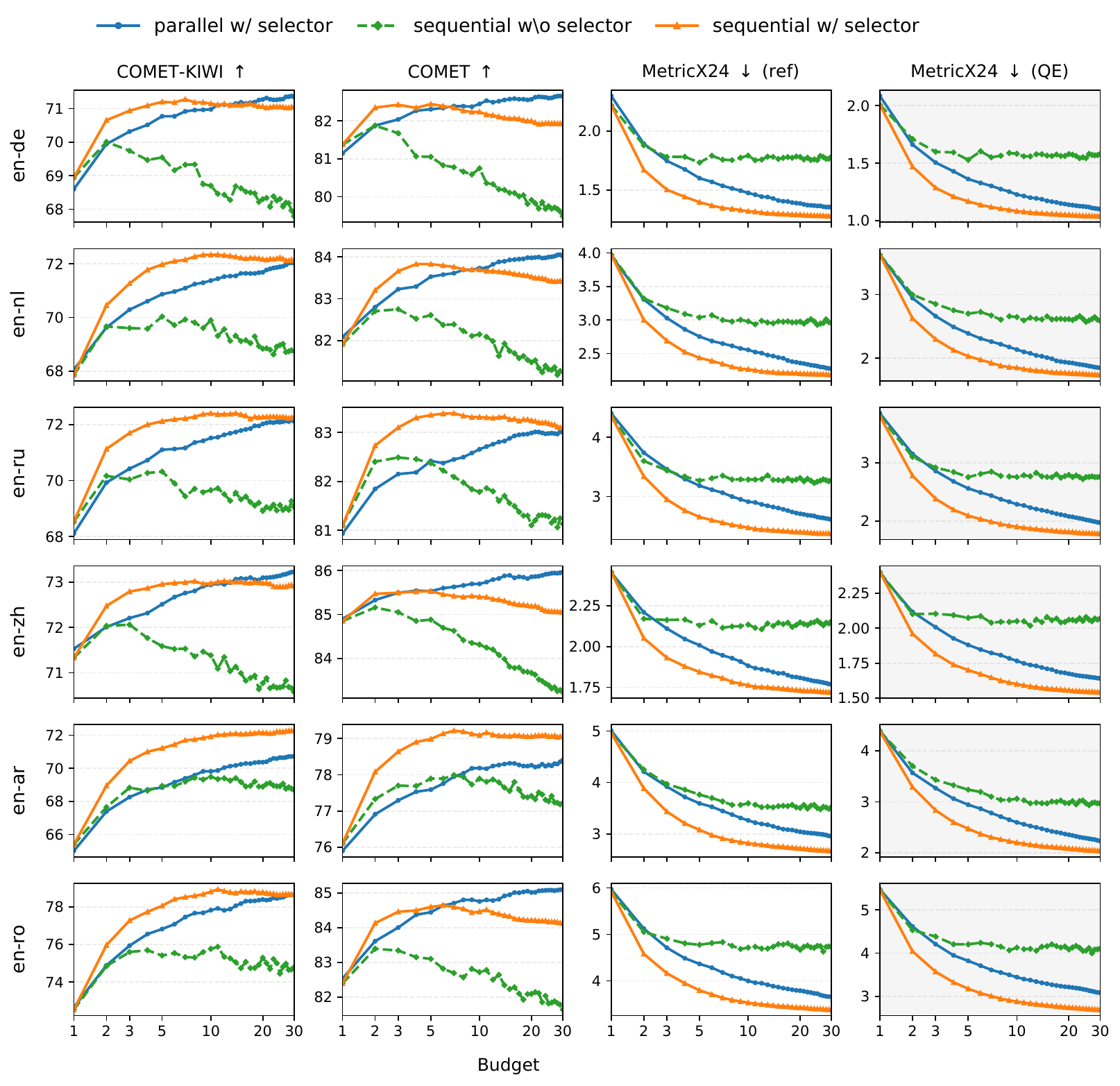}

    \caption{Quality vs.\ budget for the \texttt{Qwen-3-32B} model, when using a different selector model: MetricX24 (qe). Budget denotes the number of rounds of translation for sequential or parallel.}
    \label{fig:ablation_selector}
\end{figure*}

\Cref{fig:ablation_selector} presents a version of main results, using MetricX24 (qe) as a selector instead of \texttt{COMET-KIWI}. Results under MetricX24 (qe) as selector are consistent with the main results regarding our main finding: sequential sampling explores the output spaces more effectively, compared to parallel sampling, especially under a limited budget.

\subsection{Sampling hyperparameters \label{app:sampling_params}}
\begin{figure*}[t]
    \centering
    \setlength{\tabcolsep}{2pt}

     \includegraphics[width=1.0\textwidth]{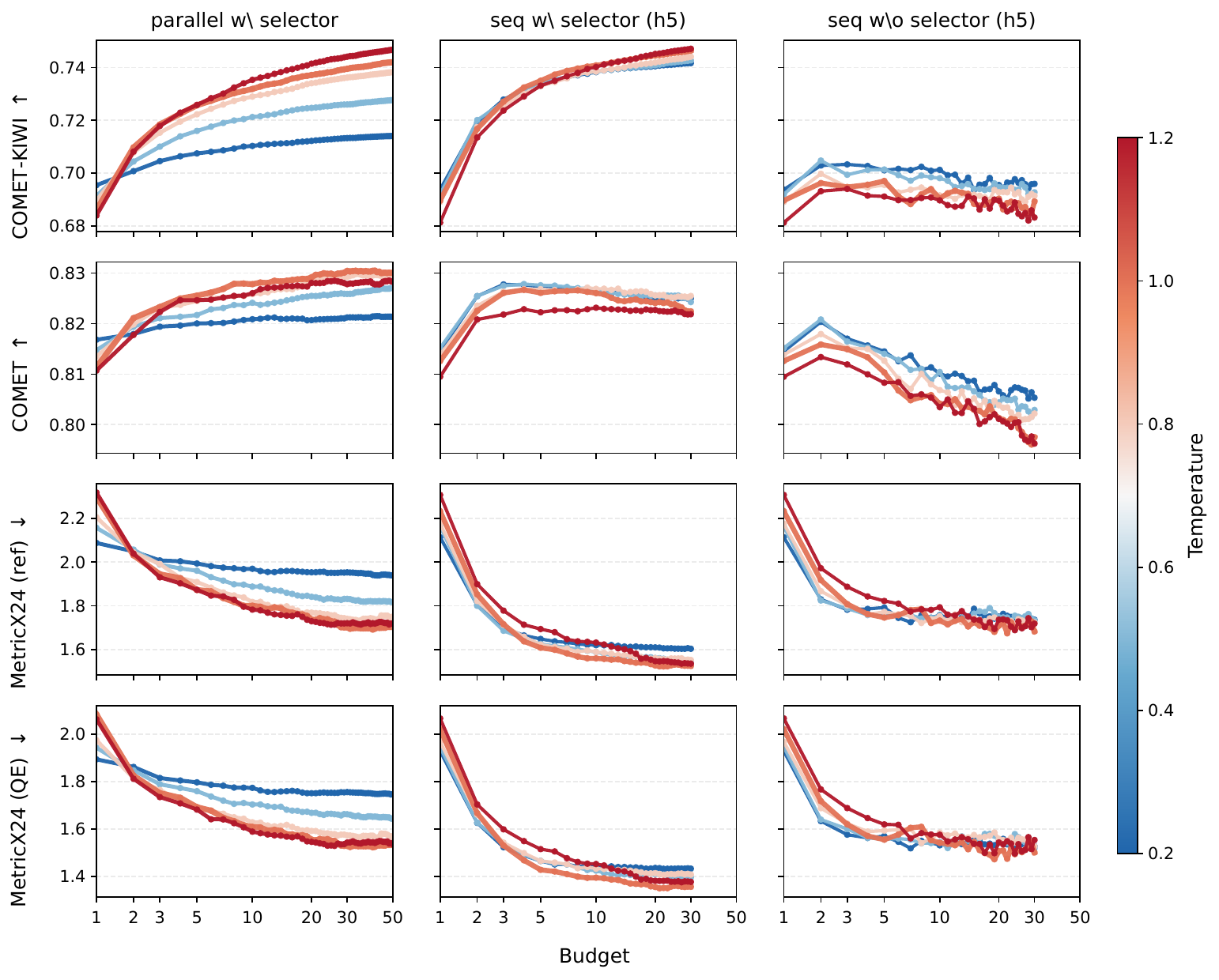}

    \caption{Temperature ablation for parallel and sequential runs for the en-de language pair. Temperature $1.0$ is near optimal for the Best-of-$N$ sampling results. For the self-refinement (no selector), lower temperature values lead to better results for the MT evaluation (high-precision, but low-diversity regime).  
    \label{fig:ablation:temp}}
\end{figure*}

We use temperature $1.0$ for all experiments, unless stated otherwise. \Cref{fig:ablation:temp} visualizes quality vs budget scaling using Best-of-$N$ sampling, showing that temperature $1.0$ is near optimal for Best-of-$N$ sampling setting. We acknowledge that without Best-of-$N$ sampling, lower temperature should be preferred to prioritize accuracy instead of diversity; however, we focus our experiments on the reranking setup due to its higher translation quality and scaling behavior.

Other sampling parameters are fixed for all experiments (top-$k$ with $k=20$, top-$p$ with $0.8$ as Qwen3 technical report recommends \citep{yang2025qwen3technicalreport}). We set the maximal number of generated tokens per turn as $4096$ for all experiments except full context experiments, where we set it to $764$ to prevent model length overflows (given a small number of collapsed extra long outputs).

\end{document}